\documentclass[journal,twoside]{IEEEtran}
\usepackage{enumitem}
\usepackage{multirow}
\usepackage {setspace}

\usepackage{mathtools}
\usepackage{cite}
\usepackage{ragged2e}
\usepackage{booktabs}
\ifCLASSINFOpdf
  \usepackage[pdftex]{graphicx}
\else
   \usepackage[dvips]{graphicx}
   \DeclareGraphicsExtensions{-eps-converted-to.pdf}
\fi
\usepackage{amsmath}
\usepackage{amsfonts}
\usepackage{amssymb}
\usepackage{amsthm}
\newtheorem{definition}{Definition} 
 
\newtheorem{theorem}{Theorem}

\newtheorem{claim}{Claim}
\newtheorem{fact}{Fact}
\theoremstyle{remark}
\newtheorem*{remark}{Remark}

\usepackage{stfloats}
\usepackage{bm}

\usepackage{algorithmic}
\usepackage{algorithm}
\usepackage{amsmath}
\allowdisplaybreaks[1]

\usepackage{array}
\ifCLASSOPTIONcompsoc
  \usepackage[caption=false,font=normalsize,labelfont=sf,textfont=sf]{subfig}
\else
  \usepackage[caption=false,font=footnotesize]{subfig}
\fi
\usepackage{color}

\usepackage{bbm}
\usepackage{url}
\begin{document}
\title{Stochastic Client Selection for Federated Learning with Volatile Clients}

% \author{Tiansheng~Huang\orcidA,~
%          Weiwei~Lin\orcidB,~
%          Li~Shen\orcidC,~
%           Keqin~Li\orcidE,~\IEEEmembership{Fellow,~IEEE},
%          and ~Albert Y. Zomaya\orcidF,~ \IEEEmembership{Fellow,~IEEE} 
 \author{Tiansheng~Huang,~
	Weiwei~Lin,~
	Li~Shen,~
	Keqin~Li,~\IEEEmembership{Fellow,~IEEE},
	and ~Albert Y. Zomaya,~ \IEEEmembership{Fellow,~IEEE} 
          % <-this % stops a space
% %\thanks{M. Shell was with the Department
% %of Electrical and Computer Engineering, Georgia Institute of Technology, Atlanta,
% %GA, 30332 USA e-mail: (see http://www.michaelshell.org/contact.html).}% <-this % stops a space
% %\thanks{J. Doe and J. Doe are with Anonymous University.}% <-this % stops a space

\thanks{This work is supported by  Key-Area Research and Development Program of Guangdong Province (2021B0101420002), National Natural Science Foundation of China (62072187, 61872084), Guangzhou Science and Technology Program key projects (202007040002), Guangzhou Development Zone Science and Technology (2020GH10), and Science and Technology Innovation 2030 –“Brain Science and Brain-like Research” Major Project (No. 2021ZD0201402 and No. 2021ZD0201405).}

\IEEEcompsocitemizethanks{\IEEEcompsocthanksitem T. Huang and W. Lin (corresponding author) are with the School of Computer Science and Engineering, South China University of Technology, Guangzhou, China. Email: tianshenghuangscut@gmail.com, linww@scut.edu.cn.\protect}

\IEEEcompsocitemizethanks{\IEEEcompsocthanksitem
L. Shen is with JD Explore Academy, Beijing, China. Email: mathshenli@gmail.com. }

\IEEEcompsocitemizethanks{\IEEEcompsocthanksitem K. Li is with the Department of Computer Science, State University of New York, New Paltz, NY 12561 USA. E-mail: lik@newpaltz.edu.
\protect}
\IEEEcompsocitemizethanks{\IEEEcompsocthanksitem A. Y. Zomaya is with the School of Computer Science, The University of Sydney, Sydney, Australia. Email: albert.zomaya@sydney.edu.au.
\protect}
}

\markboth{THIS PAPER IS UNDER REVIEW BY THE IEEE INTERNET OF THINGS JOURNAL (IoT-J)}%
{Huang \MakeLowercase{\textit{et al.}}: Stochastic Client Selection for Federated Learning with Volatile Clients}


%\IEEEtitleabstractindextext{%

%}

\maketitle

\begin{abstract}
Federated Learning (FL), arising as a privacy-preserving machine learning paradigm, has received notable attention from the public. In each round of synchronous FL training, only a fraction of available clients are chosen to participate, and the selection decision might have a significant effect on the training efficiency, as well as the final model performance. In this paper, we investigate the client selection problem under a volatile context, in which the local training of heterogeneous clients is likely to fail due to various kinds of reasons and in different levels of frequency. {\color{black}Intuitively, too much training failure might potentially reduce the training efficiency, while too much selection on clients with greater stability might introduce bias, thereby resulting in degradation of the training effectiveness.  To tackle this tradeoff, we in this paper formulate the client selection problem under joint consideration of effective participation and fairness.} Further, we propose E3CS, a stochastic client selection scheme to solve the problem, and we corroborate its effectiveness by conducting real data-based experiments. According to our experimental results, the proposed selection scheme is able to achieve up to 2x faster convergence to a fixed model accuracy while maintaining the same level of final model accuracy, compared with the state-of-the-art selection schemes.
\end{abstract}

% \fi
\IEEEpeerreviewmaketitle
\section{Introduction}
\subsection{Background}
\IEEEPARstart{D}{ata} privacy issue has received notable attention nowadays, making the acquisition of reliable and realistic data an even more challenging task. However, without the support of massive real-world data, model training by Artificial Intelligence (AI) based technique might not be realistic. In this security-demanding context, Federated Learning (FL), a privacy-preserving machine learning paradigm, has come into vision. In FL, training data possessed by a client (e.g. mobile phone, personal laptop, etc) does not need to leave the sources and be uploaded to a centralized entity for model training. All the training is done in the local devices alone, and only the post-trained models, rather than the raw data, would be exposed to other entities. By this mechanism, potential data exposure could be reduced to a minimum extent and thus making data sharing less reluctant by the data owners.
\subsection{Motivations}
\begin{figure*}[!t]
 \vspace{-1cm}
 	\centering
 	\includegraphics[ width=5in]{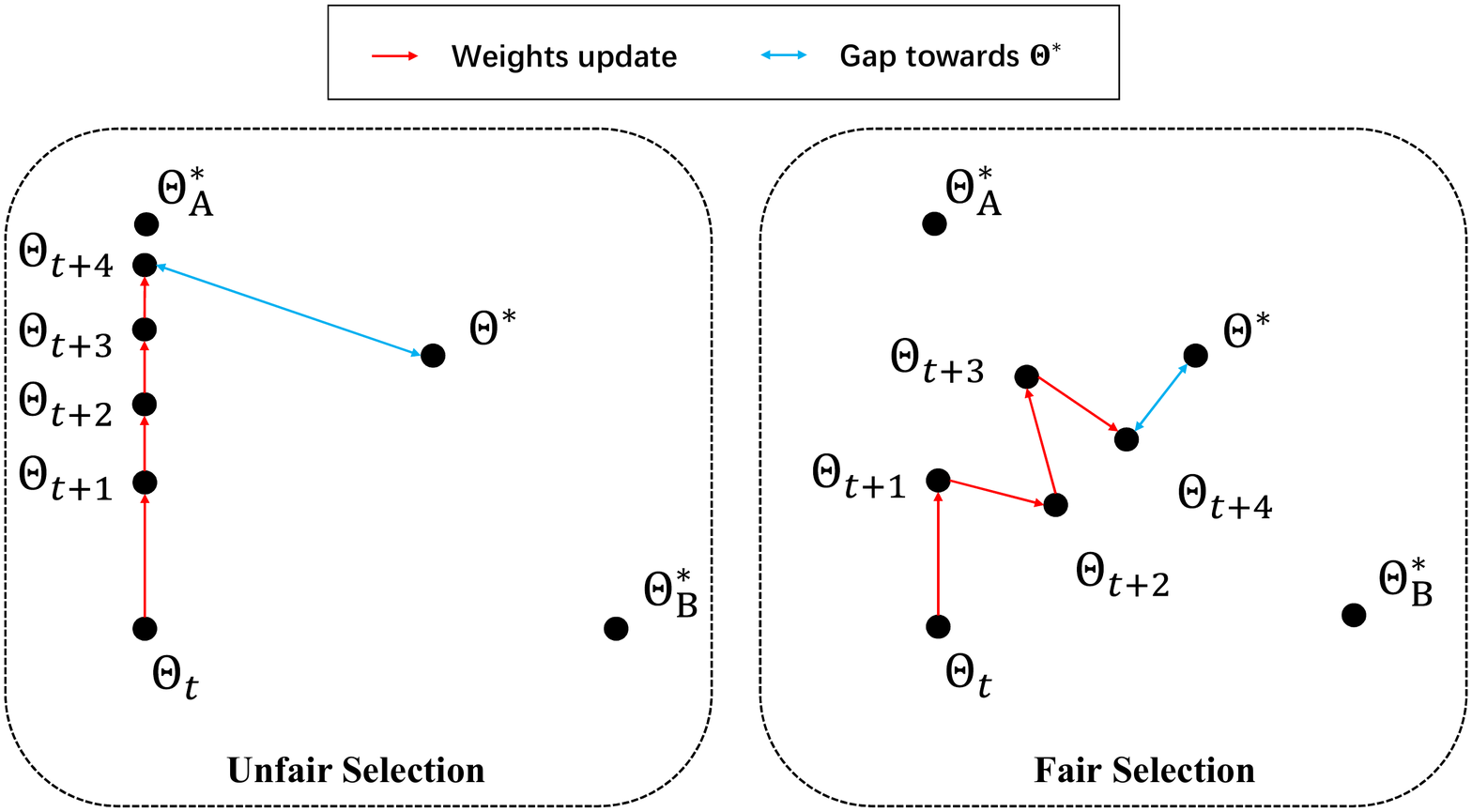}
 	% where an -eps-converted-to.pdf filename suffix will be assumed under latex, 
 	% and a .pdf suffix will be assumed for pdflatex; or what has been declared
 	% via \DeclareGraphicsExtensions.
 	\vspace{-1cm}
 	\caption{ {\color{black}An example illustrating the effect of fairness in FL, where $\bm \Theta_A^*$ and $\bm \Theta_B^*$ respectively are the local weights  of Client A and Client B (i.e., $\bm \Theta_A^*=\arg \min_{\bm \Theta}f( \bm \Theta; \mathcal{D}_A)$ and $\Theta_B^*=\arg \min_{\bm \Theta}f( \bm \Theta; \mathcal{D}_B)$ ) and $\Theta^*$ is the global optimal weights (i.e., $\Theta^*=\arg \min_{\bm \Theta}f( \bm \Theta; \mathcal{D}_A \cup \mathcal{D}_B )$). In each round of training, only one of the two clients is asked to train the global model (but no aggregation is needed in this example, which is a special case of FL). The left selection scheme consistently involves Client A while the right one gives the two clients the same opportunity to participate. It can be observed that after four rounds of training, the obtained global weights of the fairer selection is much closer to the optimal global weights $\bm \Theta^*$, so typically the model enjoys smaller loss and higher accuracy.     }    } 
 	\label{example of fairness}
 \end{figure*}

{\color{black}
	The training of canonical FL is an iterated process, in which the following basic procedure performs in sequence: 1) Server makes a client selection decision and distributes the global model to the selected participants. 2) Clients (or participants) take advantage of their local data to train the global model. Explicitly, it does multiple steps of optimization (like SGD) to the global model. 3) Clients upload the post-trained model after training is accomplished. 4) Server aggregates the uploaded model (i.e., averages the model weights of all the uploaded models) and repeats the above procedures until the aggregated model is converged.\par
	In the client selection stage, due to the limited bandwidth as well as the budget issue, not all the clients in the system are selected for training. As a result, some portion of training data is simply missing from training in each round. Under this partial participation mechanism, multiple factors, like the  \emph{ number of aggregated models} may play a critical role in the FL's convergence and performance.  Empirical observations given in \cite{ mcmahan2016communication} (i.e., the first FL paper) agree that increasing the number of participants (equivalently, enlarging the number of post-trained models to aggregate, if no training failure presents) in each round may  boost the convergence speed.      \par
	Now consider a more realistic \emph{volatile training context}, in which \emph{the selected client may not successfully return their models for aggregation in each round}. These failures could be as a result of various kinds of reasons, e.g., insufficient computing resources, user abort, and network failure, etc. This volatile context is pretty common in \emph{IoT scenario}, and typically, in this particular scenario, different clients may experience different rates of failure owing to their heterogeneous composition.  Under this more realistic consideration, and combining the observations given in \cite{ mcmahan2016communication}, we may derive a rather heuristic rule of making client selection for this context: to maximize the \emph{effective participation} in each round. That is, \emph{simply selecting the clients with the lowest failure probability (i.e., the stable ones) would be beneficial and may accelerate the training.} \par
 But this statement could be erroneous as subsequent study reveals a possible drawback of this naive 
selection scheme. The major concern arises from the violation of \emph{selection fairness}. In our previous work \cite{huang2020efficiencyboosting}, we empirically substantiate that biased selection of clients may somehow hurt the model performance (or final model accuracy). Specifically, we argue that too much selection on a specific group of clients may make the global model "drifting" towards their local optimizer. That is, the model might overfit to the data owned by those clients that we frequently select, while having poor accuracy for those seldom accessed (see Fig. \ref{example of fairness} for an example). Barring the most intuitive observation, a concurrent study \cite{cho2020client} has provided theoretical analysis on selection fairness. They present a non-vanishing bias term in the convergence error, which is exactly resulting from the \emph{selection bias} (\emph{selection skewness} in their contexts) towards some specific clients.    \par
To sum up, reasonable inference based on existing studies reveals that: 1) tendentiously selecting stable clients might increase effective participation, and thereby accelerate convergence, 2) but selection bias might deteriorate the obtained model's performance. As both the training efficiency and effectiveness are of interest in FL, we in this paper like to investigate the tradeoff between \emph{effective participation} (i.e. number of returned models for each round) and \emph{selection fairness} in a volatile training context, and discover if there is a nice way to tame these two seemingly contrastive metrics.}
	
\subsection{Contributions}
The main contributions of this paper are listed as follows:
\begin{itemize}
\item We propose a deadline-based aggregation mechanism to cope with FL aggregation in a volatile training context.
\item {\color{black} Under the new aggregation mechanism and the premise of the client's volatility, we formulate the global optimization problem for FL. To simplify the problem, we  decompose the global problem into two sub-problems based on the idea of alternating minimization. {\color{black}After that, we propose to relax the client selection sub-problem to a solvable form based on empirical observations.} }
\item We design an efficient solution termed Exp3-based Client Selection (E3CS) for the defined stochastic client selection problem. Theoretically, we derive the regret bound of E3CS, and we further discuss its relation to the canonical Exp3 in our analysis. Empirically, numerical as well as real data-based experiments are conducted to substantiate the effectiveness of our proposed solutions.
\end{itemize}
\par
{\color{black}
To the best knowledge of the authors, this is the first paper that presents a systematic study for FL under a volatile context. Before this study, almost all the literature on synchronized FL escape the potential dropout phenomenon of clients. Consequently, this paper may bring new insights into the subsequent research of synchronized FL.}
%The rest of this paper is organized as follows. Section II briefly discusses the related work. In section III, we first familiarize the readers with the basic training procedure of federated learning, and then our proposed fairness combined problem will be drafted. In Section IV, we transform the proposed intractable offline problem into an online problem and explain our estimation on the model exchange time. Then we would detail our proposed algorithm RBCS-F in the same section. After evaluating the theoretical and experimental performance of RBCS-F in Section V and Section VI, we will summarize our conclusions in Section VII.
\section{Related Works}
 Open research of FL can be classified into the following aspects.
\subsection{Statistical Heterogeneity}
 The authors in \cite{zhao2018federated} highlighted the issue of non-independently identical distribution (non-iid) of data. They presented the experimental data of FL training in a non-iid setting, which indicates a test accuracy loss of up to 51\% for CIFAR-10\footnote{\color{black}The CIFAR-10 \cite{krizhevsky2009learning} dataset is a collection of images that are commonly used to train machine learning and computer vision algorithms.} and 11\% for MNIST\footnote{\color{black}MNIST \cite{lecun1998gradient} is a collection of handwritten digits that are commonly used for training various machine learning model.}, comparing to an iid case. {\color{black} This phenomenon can be explained by statistical heterogeneity, which makes the local optimum model weights for different clients utterly different, and is proven to affect the convergence rate of FL in \cite{li2019convergence}}. As a potential solution, authors in \cite{zhao2018federated} proposed to create a small subset of public data, which is distributed to clients to form a rectified "iid" data set. A similar idea of data exchange is also available in \cite{yoshida2020hybrid,sun2020energy}. Identifying the potential privacy leakage and communication overhead of the above data-exchange-based approach, Jeong \textit{et al.} in\cite{jeong2018communication} further proposed federated augmentation (FAug), which essentially is to train a generative adversarial network (GAN) to produce the "missing" data samples, so as to make the training dataset becoming iid. However, potential privacy intrusion still persists in this data generation method, since training GAN  needs seed data samples uploaded from clients.
 \subsection{System Heterogeneity}
{\color{black}
  Another performance bottleneck faced by FL is the heterogeneous computing and network capacity of clients. In synchronous FL, aggregation could only be conducted when all the clients fully complete their local training and return their post-trained model. But the clients have utterly different performances in the real use case, which implies that some "faster" clients have to wait for the "slower", resulting in unnecessary time-consuming. \par
 To tackle this problem, Li et al. in \cite{li2018federated} allow the training epochs of different clients to be inconsistent, such that "weaker" clients can be allowed to perform less computing in each round of training. But this more flexible setting makes the original FedAvg scheme performs even worse in the case that the clients' data are \emph{highly statistically heterogeneous}. An intuitive explanation for this phenomenon is that the global model is peculiarly prone to drift (or be pulled) towards the local optimum of those clients who conducted more epochs of training, resulting in an incomplete and unbalanced global model. To cope with this arising problem, the authors in \cite{li2018federated}  proposed to use a proximal term to constrain the "distance" between the local model and global model, such that the local model of those clients with more computation (or more steps of update) would not be so drastically deviated from the global one, and therefore mitigating the drifting phenomenon. In another recent work\cite{ruan2021flexible}, the idea of imposing varying epochs for different clients is also adopted. But they propose an alternative solution for the drifting problem. Specifically, they assign "epochs" involved aggregation weights to different clients, and the clients with fewer training epochs are assigned with higher aggregation weights, such that their local model would not be overwhelmed by those with larger training epochs.\par
However, we argue that system heterogeneity in FL not only specifies the local training epochs of clients, but more aspects (such as volatility of clients) should also be considered. As such, we in this paper like to extend the notion of system heterogeneity in FL to the client's heterogeneous volatility, and discover its subsequent impact on the overall training performance. 
}
 \subsection{Client Selection}
  {\color{black} In FL, due to the communication bottleneck, often only a subset of clients could be selected to put into training in each round, which is termed as partial participation (aka partial selection). Partial participation might introduce extra bias in update gradients, thereby deteriorating FL's training performance. But the proper selection of clients in each round may potentially narrow this performance gap.}  \par
   The earliest record of this genre of research is perhaps \cite{nishio2019client},  in which Nishio \textit{et al.} proposed a rather intuitive selection scheme and highlighted the importance of client update number to the model performance. In a more recent work \cite{zeng_energy-efficient_2019}, the joint bandwidth partition and client selection issue was investigated in depth, and the joint optimization problem was solved by a rather traditional numerical method. Following this line of research, Xu \textit{et al.} in \cite{xu2020client} constructed a long-term client energy constraint to the selection problem, which essentially is to reserve energy during initial rounds of training so that more clients have chances to be involved in the later rounds of training. This study yields a very similar observation as ours, that the FL process indeed benefits if later rounds of training cover more clients. And we do believe that such a phenomenon is due to the fairness effect that we discover, i.e., the model needs more diversified data to further improve if near convergence. 
\par
In the earliest client selection study \cite{nishio2019client}, the authors assumed that the training status of clients (e.g. the training time, resource usage, etc) are known or can at least be calculated. However, client selection is not always based on a known context. Sometimes we do need some historical information (or the reputation of clients) to facilitate decision-making. In \cite{wang2020optimizing}, a reinforcement learning-based selection scheme was proposed. {\color{black}The authors take dimensionality reduced model weights as states in a Markov decision process, and use the reinforcement learning technique to optimize the selection decision for each state (i.e., each group of model weights after dimensionality reduction)}. However, this specific method needs thousands of epochs of FL data to train the reinforcement learning model and thereby may lose some of its applicability and genericity. Alternatively, an exploration and exploitation balance model, e.g., multi-arm bandit (MAB), would be more realistic. For example, Yoshida and Nishio \textit{et al.} in \cite{yoshida2020mab} leveraged a canonical MAB and further developed a rectified way to accommodate the intrinsic problem of canonical MAB, i.e., an exponential number of arms' combination. In another work\cite{xia2020multi}, Xia \textit{el al.} made use of a combinatorial MAB setting, in which the independent feedback of each arm is directly observed, enabling their designed UCB algorithm to boast a strict regret bound. Their work further took the fairness constraint into account, which appears to be a critical factor in FL training. Our previous work, \cite{huang2020efficiencyboosting} shared a very similar consideration with \cite{xia2020multi}, despite that we alternatively assumed a combinatorial contextual bandit for modeling of training time. Similarly, in \cite{lai2020oort}, an MAB-based selection was also implemented, but the feedback obtained for clients not only involved training time (or training efficiency), but the statistical efficiency (or training quality) was also jointly modeled. All of the above-mentioned work made use of a deterministic UCB-based algorithm to solve the bandit problem. In this work, rather than employing the deterministic UCB algorithm, we shall alternatively adopt a stochastic Exp3-based algorithm, which makes it more natural and intuitive to cope with the fairness constraint, without the need of imposing dynamic queues, as \cite{xia2020multi} and \cite{huang2020efficiencyboosting} did.\par
{\color{black} Selection fairness is another dominant factor in performing client selection. In our concurrent work \cite{cho2020client}, the authors have derived a theoretical analysis of the selection skewness. Based on their finding, selection bias may introduce a non-vanishing constant term in the convergence error, but bias towards clients with higher local loss may accelerate the convergence speed (see the vanishing term in their Theorem 3.1). Inspired by the derived results, they proposed a power-of-choice client selection strategy to cope with the tradeoff between the local loss of the selected clients and the selection fairness.   \par
	In \cite{cho2020client}, a stochastic selection scheme was considered in their theoretical analysis, and they concluded that selection bias towards clients with larger loss might 1) accelerate the convergence rate, but 2) enlarge the gap between the convergence value to the global optimum. However, there is still a theoretical gap between their proposed method and their theoretical result (as they did not derive the optimal sampling probability based on the result, but used a rather heuristic method to determine the extent of skewness). In another study \cite{chen2020optimal}, concrete selection probability for independent sampling has been given, in which the authors inherited and applied the basic idea from \cite{wangni2017gradient}, to make use of an unbiased estimator to do the real update. {\color{black} Specifically, authors in \cite{chen2020optimal} formulated the partial update from a client as  an unbiased estimator of the real update. Based on this formulation, their goal is to tune the selection probability to minimize the variance between partial update and the real update. They derive the optimal probability allocation to optimize the above problem. However, to derive this optimal probability allocation, real updates from all the clients have to be known in advance, which indeed violates the real intent of performing client selection -- we select a subset of clients for training in order to \emph{ save the bandwidth, as well as the computation.} Neither goal can be achieved via the optimal sampling proposed by  \cite{chen2020optimal}.    } \par
	In a high level, these two recent studies \cite{chen2020optimal} and \cite{cho2020client} actually share some common insights in terms of selection. That is, it seems to be beneficial to select those clients with a larger "distance" between the current global model and their local optimal models. In other words, it is better to select the clients whose local optimal models are most deviated from the global model, so that the global model could be pulled towards a proper direction. In \cite{chen2020optimal}, this distance is quantified as local update norm, while in \cite{cho2020client}, it is local loss. These seem to be two relevant metrics to quantify this "distance".  Also, fairness matters. Though this issue has not been formally discussed by \cite{chen2020optimal}, their unbiased estimator helps maintain some degree of fairness, since the real update of clients with less selection probability are given higher aggregation weights (see their formulation of the unbiased estimator).  However, both of these two concurrent works do not consider the volatility of clients, and its subsequent impact on selection, which is the main focus of this work. 

\subsection{Agnostic/Fair Federated Learning}
 The general average loss\footnote{Let's assume the data size of each client is the same for ease of narration.} in FL can be given by:
 \begin{equation}\text{min}_{\bm \Theta} \frac{1}{K}\sum_{i=1}^K \mathbb{E}_{(\bm x,y) \sim \mathcal{D}_i} f(\bm \Theta;(\bm x,y))
 \end{equation}
 where $\mathcal{D}=\frac{1}{K}\sum_{i=1}^K \mathcal{D}_i $ is the global data distribution ($\mathcal{D}_i$ is the local data distribution). However, in agnostic/fair federated learning, the optimization objective of FL is no longer in this form. Specifically, authors in \cite{mohri2019agnostic} argue that it is too risky to simply use this general loss as FL's optimization target, as it remains unspecified if the testing data distribution actually coincides with the average distribution among the training clients (i.e., $\mathcal{D}_{test}=\frac{1}{K}\sum_{i=1}^K \mathcal{D}_i$ may not trivially hold). A natural re-formulation of the objective is to apply a min-max principle, and change the global objective as follows: 
 \begin{equation}\text{min}_{\bm \Theta} \text{max}_{\lambda}  \mathbb{E}_{(x,y) \sim \mathcal{D}_{\bm \lambda}} f(\bm \Theta;(x,y))
 \end{equation} 
 where $\mathcal{D}_{\bm \lambda}= \sum_{i=1}^K \lambda_i \mathcal{D}_i$ is a specific kind of mixture of local data distribution. By this transformation, the goal of optimization has transferred to find optimal weights $\bm \Theta$ which minimizes the expected empirical loss under \emph{any} possible mixture of local data distribution (i.e., agnostic distribution). \par
However, agnostic federated learning applying the min-max principle might be too rigid, as it maximizes the model's worst performance on any mixture of training data. Li et al. in \cite{li2019fair} proposed to relax the problem  by changing the optimization objective to: 
\begin{equation}\text{min}_{\bm \Theta}  \frac{1}{K}\sum_{i=1}^K \mathbb{E}_{(x,y) \sim \mathcal{D}_{i}} \frac{1}{q+1}f^{q+1}(\bm \Theta;(x,y))
\end{equation} 
where $q$ is the fairness parameter. By this transformation, the authors introduce \emph{good-intent fairness}, making the goal is instead to ensure that the training procedure does not overfit a model to any one device at the expense of another. Or more concretely, they like to acquire a global model that fits local data from each client, thereby reducing the variance of testing accuracy for each client. This new design is particularly fit to the scenario that the obtained model is used by clients themselves (but not deployed elsewhere). Obviously, fairness has to be nicely maintained in this case, since clients have no reasons to participate if the acquired model has poor performance on their local data. Moreover, this objective is indeed more flexible than the agnostic min-max objective, in that the fairness extent can be adjusted by tuning $q$, and therefore should be more applicable in some use cases of federated learning.\par
However, in this paper we stick to the optimization of general federated learning, in which the optimization is still the general weighted average loss. And please note that the notion of fairness (i.e., good-intent fairness) discussed in this subsection and the selection fairness we mention in this paper are not exactly the same notion, since the objectives of maintaining them are different. For good-intent fairness, the goal of maintaining it is to ensure the training loss for different clients to be more uniform. For the selection fairness we discuss in this paper, it is maintained in order to minimize the average loss (consider that bias selection might introduce difficulty in reducing average loss), but we don't consider if the local empirical losses are uniform or not. 
}
\section{A deadline-based aggregation mechanism}
\begin{figure*}[!t]
\centering
\includegraphics[ width=6.2in]{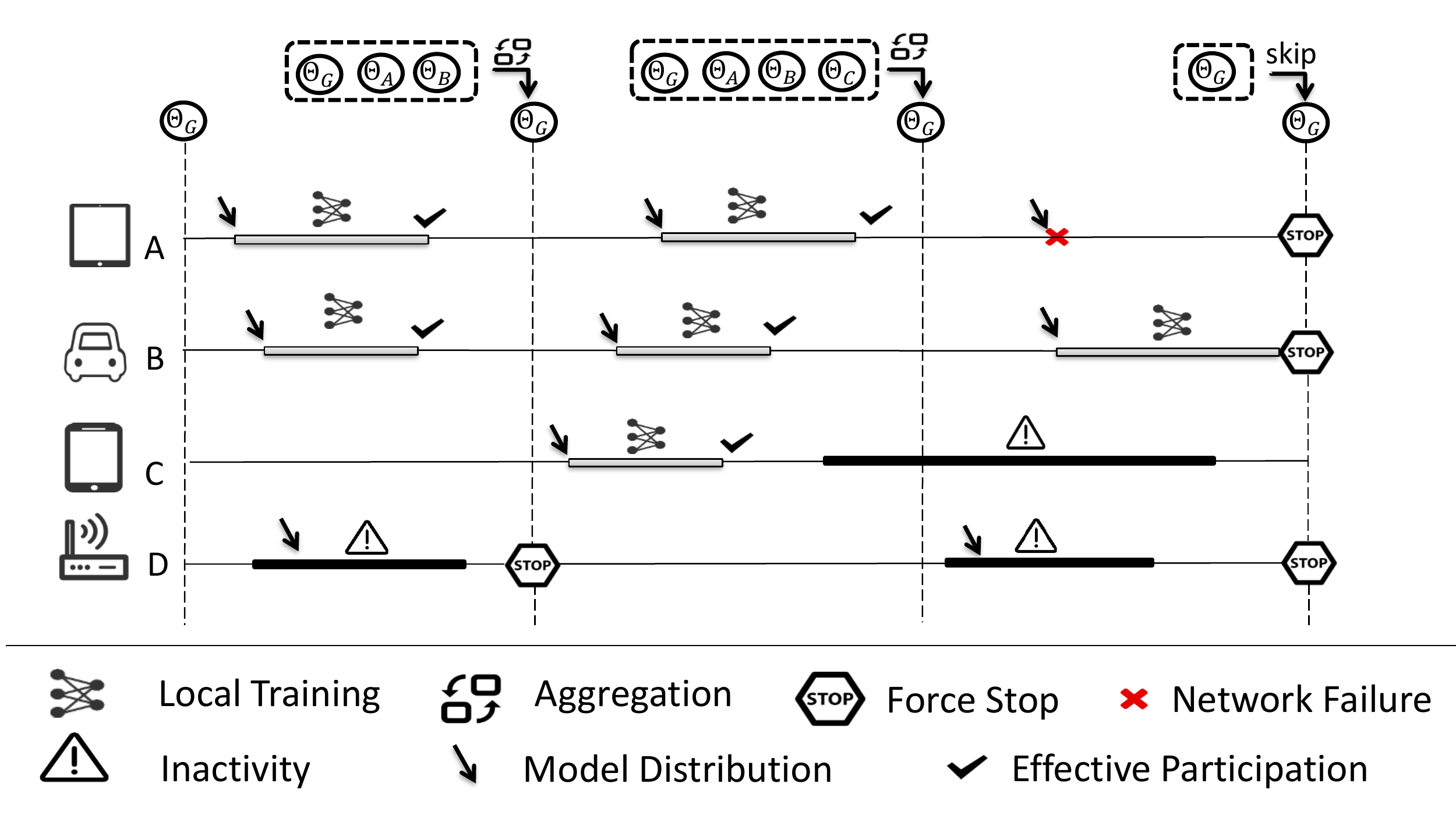}
% where an -eps-converted-to.pdf filename suffix will be assumed under latex, 
% and a .pdf suffix will be assumed for pdflatex; or what has been declared
% via \DeclareGraphicsExtensions.
\vspace{-0.2cm}
\caption{Illustration of deadline-based aggregation mechanism of FL. $\bm \Theta_G$, $\bm \Theta_A$, $\bm \Theta_B$, $\bm \Theta_C$, $\bm \Theta_D$ presented in the figure respectively correspond to the global model, and the post-trained model of client A, B, C and D.  In this example, three of four clients are chosen to participate in each round. In the first round, the server selects clients A, B, D to participate, and then distributes global model $\bm \Theta_G$ to them. But during the first round of training, client D dropouts as a result of a system crash, so only the models from clients A and B are returned before the aggregation deadline, and both of them are counted as effective participation. As no models are returned from client D, a force stop command is issued to her. After aggregation in server, a new global model is produced, {\color{black} via weighted averaging like $\bm \Theta_G= \omega_A \bm \Theta_A+ \omega_B \bm \Theta_B+ (1-\omega_A-\omega_B)\bm \Theta_G$, where $\omega_A $ and $\omega_B$ are the aggregation weights of a client (see Eq. (\ref{ml problem}) for their definition).  }  The aggregated model then substitutes the old global model. Then the second round of training ensues, in which clients A, B, and C are chosen to participate. This time, all of the three return their models on time, so no force stop command needs to be issued. In the third round of training, clients A, B, and D are chosen again, but unfortunately, no effective participation is produced in this round. Client A suffers from network failure, making the model lost during transmission. Client B cannot fulfill the designated training epochs before the deadline, and client D  crashes again during training. Then, force stop commands are issued to all the three chosen clients. Obviously, this round of training is futile, as the global model is not updated and remains the same as that in the previous round.       }
\label{system FL}
\end{figure*}
To cope with the potential drop-out of clients, we adopt in this paper a deadline-based aggregation mechanism, in which aggregation is made once a fixed deadline is met, such that the issue of "perpetual waiting" could be nicely prevented. To be specific, we involve the following stages in sequence during one round of FL training.
\par
 \textbf{Client selection and model distribution.} In this stage, the server determines participants over the available clients and correspondingly distributes the global model to them. After distribution, the main process of the server would sleep and wait until a fixed deadline is met.
\par
 \textbf{Local training.} In this stage, the selected clients conduct designated epochs of training with a local optimizer (e.g., SGD) on the basis of its local data and the distributed global model. {\color{black}The computing epochs of different clients could be designated to different values in advance, as clients may have different computing capacities (i.e., system heterogeneity) }.
\par
\textbf{Model transmission:} Once the local training of the clients completes successfully, the post-trained model would be transmitted to the aggregation server immediately.
\par
 \textbf{Force stop.} Once the deadline is met, the server issues the "Force Stop" command to the selected clients, after hearing which, clients that are still on the stage of training or model transmission will stop and abort immediately. Models returned after the deadline will be dropped. 
\par
\textbf{Aggregation.} The server will check the successful returned model of their validity, after which, aggregation based on FedAvg \cite{mcmahan2016communication} (or possibly other aggregation schemes) is conducted. After this stage, the server would repeat the above procedures to start a new training round.

For a vivid understanding of the training process, we refer the readers to Fig. \ref{system FL}, in which an example of three rounds of FL training  is demonstrated.
%In this paper, we consider an edge-coordinated federated learning system, in which edge is functioning as a model aggregator, and the clients (mostly mobile devices) are responsible for doing local training over their private data on behalf of the model's owner. We adopt in our system the most-accepted synchronous scheme for federated learning, which is characterized by training in iterations. For clearness, now we will explicitly explain the workflow of our synchronized scheme by giving four sequential stages of training, as follows:
%\begin{enumerate}
%\item At the very beginning of a new iteration, the clients first report their willingness to participate in the training as well as a few client-side information, which will be used for the client selection in the next stage.
%\item In the second step, the scheduler conducts client selection to choose a portion of participants among the volunteers in light of the provided information.
%\item Global model is distributed to the selected clients. After receiving the model, the clients conduct local training using their private data and update their local model. Once the training finished, the local model will be returned to the MEC server. The time span of this round is known as ''model exchange time''.
%\item The collected local models are aggregated by the server, substituting the original global model that once being distributed, and then it proceeds to step 1) to start a new iteration.
%\end{enumerate}
%For a more vivid presentation of the training process, we refer the readers to Fig. \ref{system FL}.
\section{Problem Formulation}
Optimization over federated learning under the deadline-based aggregation mechanism is characterized by a tuple $(\mathcal{K}, \{x_{i,t}\}_{i \in \mathcal{K},t\in \mathcal{T} }, o_1(\cdot), \{A_t\}_{ t \in \mathcal{T} }$). Here, the set $\mathcal{K}\triangleq \{1,2, \dots, K\}$ represents the total number of  $K$ accessible clients in the system,  the training round is characterized by $t \in \mathcal{T} \triangleq \{1, \dots,T\}$, $\{x_{i,t}\}_{i \in \mathcal{K},t\in \mathcal{T} }$ is the success status of a client in each round,  $o_1$ is the local update operation, and $\{A_t\}_{ t \in \mathcal{T} }$ is a set that captures the selected clients each round. We would specify how these components constitute our optimization problem in the following. 
{\color{black}
\subsection{Basic Assumption and Global Problem} \label{deadline model}
In this subsection, we shall illustrate the basic setting of federated learning with volatile clients, and the global optimization problem.
\par
\textbf{Client Dropout.}
 There is a chance for the clients to drop-out in the middle of training as a result of technical failures (or the client proactively quits the training process). This dropout phenomenon is quite common in IoT scenarios. To model this critical concern, we introduce a binary number i.e., $x_{i,t}$, to capture the status of client $i$ in round $t$. Formally, $x_{i,t}=1$ means the training of client $i$ would succeed, and vice versa. 
\par
\textbf{Cardinality Constraint.}
 In each round of FL,  only a fraction of clients could be selected to participate due to limited bandwidth (the same setting as in \cite{wu2020safa,wu2020accelerating, huang2020efficiencyboosting}). Now we assume that we select $k$ out of $K$ clients in each round, formally we need to ensure:
 \begin{equation}
 	\label{cardinality constraint1}
 	|A_t|= k \quad t\in \mathcal{T} 
 \end{equation}
 where $A_t$ is the selection set in round $t$, and $|A_t|$ is the cardinality of the selection set.
\par
  \textbf{Global Problem.}
The global optimization problem that we like to solve within the total time frame $T$ is as below:
\begin{equation}
		\begin{split}
\text{P1}: \quad &\min _{o_1, \{A_t\}_{t \in \mathcal{T}} }  \sum_{i=1}^{K}  \mathbb{E}_{(\bm x,y) \sim \mathcal{D}_{i}} \omega_i f\left( \bm \Theta_{T+1}  ; (\bm x,y)\right) \\
		\text{s.t.} \quad &	{\color{black}	|A_t|= k \quad  }  \qquad\text{(cardinality constraint)} \\
		& \bm \Theta_{i,t}= o_1 ( \bm \Theta_{t}, \mathcal{D}_i   ) \quad  \text{(local update operation)}\\
		& {\color{black} \bm \Theta_{t+1} =  \sum_{i\in A_t\text{ and  }x_{i,t} \neq 0} \omega_i \bm  \Theta_{i,t} +\sum_{i \notin A_t\text{ or }x_{i,t} = 0} \omega_i  \bm \Theta_{t} }  \\
		& \qquad \qquad    \text{(aggregation operation for volatile context)}\\
	\end{split}
\label{ml problem}
\end{equation}
where, 
\begin{itemize}
\item $\mathcal{D}_i$ is used to denote the local data distribution of client $i$ and $(\bm x,y)$ specifies one piece of data (with $\bm x$ being input and $y$ being label). 
\item $\bm \Theta_{t} \in \mathbb{R}^d $ and $\bm \Theta_{i,t} \in \mathbb{R}^d$  are the global model weights and the local model weights after  local update (i.e., updated weights under the premise that local update has been successfully accomplished). 
\item $\omega_i=|D_i| /\sum_{i \in [K]}|D_i|$ is used to denote the proportion of data of a specific client, where $|D_i|$ is the number of data the $i$-th client hosts. 
\item $f(\bm \Theta_{T+1};\cdot)$ is the empirical loss of a piece of data given  model weights $\bm \Theta_{T+1}$. Our primary objective is to minimize the average empirical loss over data residing in all the clients.  
\item $o_1(\cdot)$ is the local update operation. Different schemes in the literature apply different update techniques here. For example, FedAvg\cite{mcmahan2016communication} applies stochastic gradient descent (SGD) over the general loss function here.  In another work, FedProx \cite{li2018federated} employs SGD over a proximal term-involved loss function during the update.   
\end{itemize}
\textit{P1} is a comprehensive global problem for FL in a volatile context. It involves the optimization of two key processes of FL, i.e., the local update scheme ($o_1$), and the client selection scheme ($A_t$). The volatile context that we consider in this paper is directly reflected by the aggregation operation shown in the third constraint.  {\color{black} In order to accommodate the volatile training context, the aggregation operation is slightly modified from its original form of FedAvg (original formulation is shown in Eq. (3) in \cite{fraboni2021clustered}). In our modified version, the updates from clients with unsuccessful participation (either not selected or failed in the middle of training) are replaced by the current global model, while in its original form, only those not selected are replaced.   } \par 
 \subsection{Alternative Optimization} The global problem P1 is a combinatorial optimization problem, in which multiple variables are coupling together, imposing great challenges for optimization. To lower the complexity,  we adopt the idea of alternating minimization to decompose the problem. Specifically, we divide the variables into two blocks, namely, the client selection block (i.e., $A_t$) and the local operation block (i.e., $o_1$). By the decomposition, we are allowed to split the problem into two sub-problems, namely, 
 \begin{equation}
 	\begin{split}
 			&\text{P1-SUB1:} \quad  \min _{o_1}  \sum_{i=1}^{K}  \mathbb{E}_{(\bm x,y) \sim \mathcal{D}_{i}} \omega_i f\left( \bm \Theta_{T+1}  ; (\bm x,y)\right) \\
 		&\text{P1-SUB2:} \quad \min _{\{A_t\}_{t \in \mathcal{T}} } \sum_{i=1}^{K}  \mathbb{E}_{(\bm x,y) \sim \mathcal{D}_{i}} \omega_i  f\left( \bm \Theta_{T+1}  ; (\bm x,y)\right) 
		\\ &\text{s.t.  same constraints with P1}
	\end{split}
\end{equation}
P1-SUB1 is a standard optimization problem for FL, to which multiple existing studies (e.g. \cite{li2018federated,mcmahan2016communication,ruan2021flexible}) have given concrete solutions, in order to cope with two notorious issues in FL (i.e., system heterogeneity and statistical heterogeneity).   However, it remains unexplored how to solve the client selection sub-problem P1-SUB2 in the presence of our third constraint under volatile context. In this paper, we shall focus on the solution of P1-SUB2, while fixing existing solutions for P1-SUB1 to alternatively optimize the problem.  

\subsection{Stochastic Optimization of Client Selection Sub-problem}
{\color{black} Insipred by \cite{li2018federated}, we consider to use multinomial sampling \footnote{ {\color{black} Drawing samples from a multinomial distribution with multiple trials is equivalent to draw values from a set, say, $\{1,2,\dots,K\}$ in different probability and for multiple times  (quite like rolling for multiple times a special dice, which has unequal probability for each side).  And no replacement specifies that the drawn values between different trials cannot be the same (which reflects the basic property of a combination).  A simple implementation (or simulation) of this multinomial distribution can be found in Section 5 of a draft textbook (can be accessed through http://www.stat.cmu.edu/~cshalizi/ADAfaEPoV/ ) } } to stochastically determine the selection combination each round. Specifically, we assume: (i) $A_t \sim \text{multinomialNR}( \bm p_t/k, k )  \quad \forall{t} \in \mathcal{T}$, where \text{multinomialNR}($\cdot$) denotes a multinomial distribution without replacement, (ii) $\sum_{i=1}^K p_{i,t} = k $, and (iii) $0 \leq p_{i,t}\leq 1$. These assumptions suffice to ensure that $|A_t|= k$ and $\mathbb{E}[\mathbb{I}_{\{i \in A_t  \}}]=p_{i,t}$, which means the probability that a client being selected  is $p_{i,t}$.    By this transformation, we obtain a stochastic version of P1-SUB2, as follows:
	 \begin{equation}
		\begin{split}
	& \text{P2:} \min _{\{\bm p_t\}_{t \in \mathcal{T}} } \sum_{i=1}^{K} \mathbb{E}_{(\bm x,y) \sim \mathcal{D}_{i}} \omega_i f\left( \bm \Theta_{T+1}  ; (\bm x,y)\right) \\
	 \text{s.t.} \qquad  	&	A_t \sim \text{multinomialNR}( \bm p_t/k, k )  \quad  \\
	 & \sum_{i=1 }^K p_{i,t}= k  \quad  \\ & 0\leq p_{i,t} \leq 1 \quad  \\
	& \bm \Theta_{i,t}= o_1 ( \bm \Theta_{t}, \mathcal{D}_i   )  \\
	& {\color{black} \bm \Theta_{t+1} =  \sum_{i\in A_t\text{ and  }x_{i,t} \neq 0} \omega_i \bm  \Theta_{i,t} +\sum_{i \notin A_t\text{ or }x_{i,t} = 0} \omega_i  \bm \Theta_{t} }  \\
\end{split}
\end{equation}
	
}

{\color{black}By this transformation, the problem's optimization target has transferred to the probability allocation $\bm p_t \triangleq (p_{1,t}, \dots,p_{K,t})$. However, after transformation, it is still much complicated to derive a uniform solution to problem P2, given that the local update operation $o_1$ could be diversified, and there is not a universal method to quantify their impacts on the client selection policy.} 

{\color{black}
\subsection{Problem Relaxation based on Empirical Findings}
Then we shall shed light on how we resort to an empirical observation to further relax the problem. More specifically, under the framework of both FedAvg and FedProx (i.e., fixing $o_1$ according to their setting), we find empirically that two critical factors, i.e., \emph{expected cumulative effective participation} and \emph{selection fairness}, might impose a critical effect on the optimization of $\bm p_t$ in the presence of volatility. Now we shall introduce these two factors in sequence, as follows:
}
}
\par
\textbf{ Expected Cumulative Effective Participation.}
Factually, the training processes of volatile clients are not always bound to succeed. In the course of training, there are various kinds of reasons for the clients to drop-out (or crash), i.e., not capable of returning models to the server on time. For example, clients might unintentionally shut down the training process due to resource limitations. Or perhaps clients are too slow to return the post-trained model, leading to a futile participation \footnote{In our deadline-based aggregation scheme, FL server would set up a deadline for collecting models from clients and model submitted after which would be dropped. }. From the perspective of an FL server, a frequent failure of the clients is the least desirable to see (since there are fewer models available for aggregation), and therefore, those "stable" clients should be more welcome than the "volatile" ones. {\color{black} Also, empirical observation given in the first FL paper \cite{mcmahan2016communication} indicates that increasing the client fraction ($C$ in their context) would accelerate the learning process\footnote{\color{black}See the conclusion made by Table 1 in \cite{mcmahan2016communication}. }, which means increasing the number of aggregated models would be beneficial in our volatile context.  } 
To model this particular concern, we define the metric named \textit{expected Cumulative Effective Participation} (CEP), i.e., the expected number of post-trained models that have been successfully returned. {\color{black} Formally, we propose to  maximize the expected CEP as follows:
\begin{equation}
	\mathbb{E} \left [ \sum_{t=1}^{T} \sum_{i =1}^K \mathbb{I}_{\{ i \in A_t \}} x_{i,t} \right ],
\end{equation}
where $\mathbb{I}_{\{ i \in A_t \}} x_{i,t}$ equals to one only if the i-th client is selected (i.e., $i \in A_t $) and the training is success (i.e., $x_{i,t}=1$). Further consider the fact that $\mathbb{E}[\mathbb{I}_{\{i \in A_t  \}}]=p_{i,t}$, we obtain that:
 \begin{equation}
 	\mathbb{E} \left [ \sum_{t=1}^{T} \sum_{i =1}^K \mathbb{I}_{\{ i \in A_t \}} x_{i,t} \right ] = \sum_{t=1}^{T} \sum_{i =1}^K p_{i,t} x_{i,t}.
 \end{equation}.
 
} {\color{black} While maximizing expected CEP, conceivably, the model would be biased towards data from clients that have higher success rate. To control the extent of skewness, we introduce the fairness metric in the following. }
\par
\textbf{ Selection Fairness.}
\label{Fairness constraint}
In the real training scenario, data that resides in clients are normally statistically heterogeneous and therefore have its own value in promoting training performance. If we intentionally skip the training of some clients (perhaps those with a higher probability to drop-out), then most likely, the final model accuracy would suffer an undesirable loss, since the global model will lean towards the local optimal models of those frequently selected. {\color{black}This training pattern is akin to consistently using a portion of data for training in pure centralized machine learning. Conceivably, with this biased pattern, the trained model will be overfitted to a subset of the frequently involved data, thereby degrading model accuracy. {\color{black}Empirical observation of this performance degradation phenomenon is reported in \cite{huang2020efficiencyboosting},  and is theoretically studied in \cite{cho2020client}.}   \par

As such, we need to ensure fairness to control (though not eliminate) the selection bias. We achieve this goal by reserving some probability for each client to be selected.} More concretely, we need to make sure the probability that each client being selected should at least be $\sigma_t$, such that each client is at least given some chances to get involved. Formally, we need to make sure:
\begin{equation}
\label{group fairness}
 p_{i,t} \geq \sigma_t  \quad \forall i \in \mathcal{K},  \forall t \in \mathcal{T}, 
 \end{equation}
Obviously, we can find that the higher the value of $\sigma_t$ is, the evener the selection would become. In this regard,  we refer to $\sigma_t$ as \textit{fairness quota} \footnote{It is noticeable that we allow the fairness quota in each round not necessarily be the same. This setting is associated with some intrinsic features of FL training, which we will detail in our experimental part.}, which directly measures the fairness degree of this round of selection.
{\color{black}
Also, we need to ensure the constraint $0\leq \sigma_t \leq k/K$ holds for any $t$, since i) setting $\sigma_t <0$ is meaningless under the assumption $p_{i,t} \geq 0$, and (ii) setting $\sigma_t>k/K$ violates the assumption $\sum_{i=1}^K p_{i,t}=k$. Importantly, $\sigma_t=k/K$ implies that $p_{i,t}= k/K$ holds for all clients. In this case,  all the clients share the same probability to be selected, and selection would be reduced to  uniform sampling adopted by FedAvg. Therefore, the fairness quota $\sigma_t$ serves as a key hyper-parameter to tradeoff selection skewness and training efficiency (which could be promoted via biased selection over the clients with higher success rate). }
\par 
 \textbf{Relaxed Client Selection Problem.}
Putting all the pieces together, now we shall introduce the relaxed client selection problem, as follows:
 {\color{black}
\begin{equation}
	\begin{split}
		\text{ P3:}  &\quad \max \limits_{\{\bm p_{t}\}} \quad   \sum_{t=1}^{T} \sum_{i=1 }^K p_{i,t} x_{i,t} \\
		\text{s.t.} & \quad \sum_{i=1 }^K p_{i,t}= k \quad \forall{t} \in \mathcal{T}  \\ 
		&  \quad  \sigma_t \leq p_{i,t} \leq 1 \quad  \forall{t} \in \mathcal{T}, \forall i \in \mathcal{K} \\
	\end{split}
\end{equation} \par  
}
{\color{black}
However, the problem cannot be solved offline, since the success status of a client, i.e.,  $x_{i,t}$ is generally unknown by the coordinator before round $t$. We in the following section would introduce an adversary bandit-based online scheduling solution for problem-solving. }
\section{Solution and algorithms}
In this section, we shall give a solution about how to determine the selection probability of different clients such that the expected cumulative effective participation is being maximized in P3. Before our formal introduction, we shall first give a preliminary illustration of Exp3, a rather intuitive solution of the adversarial bandit problem, which serves as the base of our proposed solution.
\subsection{Preliminary Introduction to Canonical Exp3}
{\color{black}
 Subjecting to the selection constraint $0 \leq p_{i,t} \leq 1$ and $\sum_{i} p_{i,t}=1$, the objective of adversary bandit is to maximize the following objective:
$\max \limits_{\{\bm p_{t}\}} \quad   \sum_{t=1}^{T} \sum_{i} p_{i,t} x_{i,t} $ where $x_{i,t}$ is the reward of drawing the $i$-th arm, and  $p_{i,t}$ is the selection probability.

As a solution to adversary bandit, Exp3  first defines  the \textit{unbiased estimator} of the real outcome of arms, as follows: }
\begin{equation}
	{\color{black}
\hat{x}_{i,t}=\frac{\mathbb{I}_{ \{ i \in A_t\} }}{p_{i,t}} x_{i,t}}
\end{equation}
It can be found that $\mathbb{E}[\hat{x}_{i,t}]=x_{i,t}$. Using the unbiased estimator, the \textit{exponential weights}, which reveals the future rewards of an arm, can be formulated as follows:
\begin{equation}
w_{i,t+1}= w_{i,t} \exp\left(\eta \hat{x}_{i,t}   \right )  
 \end{equation}
As the obtained weights reflect our estimation of each arm's potential outcome, intuitively, we might need to allocate more selection probability to those with a higher weight. Moreover, we need to ensure that the allocated probability sums up to 1. An intuitive way to achieve this requirement is to calculate the allocated probability as follows: 
\begin{equation}
p_{i,t}= \frac{w_{i,t}}{\sum_{j \in \mathcal{K} } w_{j,t} }
 \end{equation}
{\color{black}By this kind of probability allocation, the algorithm should be able to identify the arm with higher rewards and allocate them more probability.} However, the canonical Exp3 could only apply to the situation that only one arm is selected in each round. Moreover, the canonical Exp3 cannot guarantee that the selection probability of each arm is at least greater than a constant. Therefore, we need to make necessary adaptations towards it in order to enable multiple plays each round, and accommodate the fairness constraint in our context,  In the next subsection, we shall introduce our adaptation based on an elementary framework \cite{uchiya2010algorithms} for adversarial bandit with multiple plays.

\subsection{EXP3 with Multiple Play and Fairness Constraint}
%\begin{assumption}
%\label{additional assumption}
%If there exists group $n$ and $n^{\prime}$ such that $K_n \neq K_{n^{\prime}}$, the setting of group fairness $\sigma$ should satisfy:
%\begin{equation}
%\sigma \leq \frac{K_{min}(K-k)}{K-K_{min}N}
%\end{equation}
%where $K_{min}=\min_{n \in \mathcal{N}}\{ K_n\}$
%\end{assumption}
%The assumption is necessary to prevent overflowed probability allocation, which is a serious issue that plagues our proposed solution under the unbalanced grouping situation. However, we need to note that this additional assumption would only be effective in some really rare setting (almost impossible in FL client selection setting). In most the practical setting of FL client selection, our solution serves well under the basic assumption $\sigma< k/n$ (see section \ref{Grouping and Fairness}). The detail of why we need this particular assumption would be specified later. \par
{\color{black}
As can be found in problem P3, our formulated problem is in the same form of adversary bandit problem with multiple plays, though with an extra constraint over the minimum probability to be allocated to a client. In the following, we propose an Exp3-based solution for problem-solving.}
\par
\textbf{Unbiased Estimator and Exponential Weights.}
 We consistently use the same unbiased estimator of $x_{i,t}$ as in the canonical Exp3 solution, namely:
{\color{black}
\begin{equation}
\label{unbised estimation final}
\hat{x}_{i,t}=\frac{\mathbb{I}_{ \{ i \in A_t\} }}{p_{i,t}}x_{i,t}
\end{equation}
Intuitively, we can derive that $\hat{x}_{i,t} \in ( 0,\infty] $ and $\mathbb{E}[\hat{x}_{i,t}  ]= x_{i,t}$. }\par
But the exponential weights update should be modified to:
\begin{equation}
\label{weight update final}
w_{i,t+1}= \begin{cases} w_{i,t} \exp\left(\frac{(k-K\sigma_{t})\eta \hat{x}_{i,t}  }{K} \right )  &  i \notin S_t \\ w_{i,t} &  i \in S_t   \end{cases}
 \end{equation}
 where $0<\eta<1$ denotes the learning rate of weights update. $S_t$ is the set of clients that experience probability overflow during the probability allocation stages, which will be specified later. The in-depth reason of this modification is available in our proof of regret (see Appendix \ref{regret proof}).  
\par
\textbf{Probability Allocation.}
\label{probability allocation section}
Given the exponential weights of clients, the way we derive probability allocation should be revised to a more sophisticated form, in order to accommodate $ p_{i,t} \geq \sigma_t$ (i.e., the expected fairness constraint) and $\sum_{i=1}^K p_{i,t} =k$.  In order to qualify these two constraints, our idea is to first allocate a total amount of $\sigma_t$ probability to each client to accommodate the fairness constraint, and then further allocate the residual amount of probability (i.e., $k-K\sigma_t$, since $\sum_{i=1}^K p_{i,t} =k$, which is imposed by the cardinality constraint) according to the clients' weight. Formally, we give:
\begin{equation}
\label{temp inter group selection probability}
p_{i,t}= \sigma_t+ (k- K\sigma_t)\frac{ w_{i,t}}{\sum_{j \in \mathcal{K}} w_{j,t} }
\end{equation}
 However,  a critical issue might persist in this form of probability allocation: the allocated $p_{i,t}$ would be possibly greater than 1 if the exponential weight of a client is too large, which violates the constraint $p_{i,t} \leq 1 $ in P3. We refer to such an unexpected phenomenon as \textit{probability overflow}. 
 
  \textbf{Capping the probability.} To address the probability overflow issue, we shall employ a capping-based technique for the probability allocation.  \par

Formally, we need to rewrite Eq. (\ref{temp inter group selection probability}) as follows:
\begin{equation}
\label{inter group selection probability}
p_{i,t}= \sigma_t+ (k- K\sigma_t) \frac{w_{i,t}^{\prime}}{\sum_{j\in \mathcal{K}} w_{j,t}^{\prime} }
\end{equation}
where $w_{i,t}^{\prime}=\min \left\{ w_{i,t}, (1-\sigma_t) \alpha_t \right\}$.\par

By introducing the revised weights, we cap those weights that are too large to a smaller value, i.e., $\alpha_t$, such that the selection probability of those ”overflowed“ clients would be capped to 1.  Now the problem has been transferred to choose a proper $\alpha_t  $, making $p_{i,t} \leq 1 $ holds for all $i$. But before we introduce how to derive $\alpha_t  $, we shall first determine the existence of at least one qualified $\alpha_t$ to ensure the applicability of our capping method. Here we need the following claim:
\begin{claim}
\label{existence of inter alpha_t}
By the calculation indicated by Eq. (\ref{inter group selection probability}), there exists at least one qualified $\alpha_t$ such that $ p_{i,t} \leq  1$ holds for all $i \in \mathcal{K}$.
\end{claim} 
\begin{proof}
To formally start our proof, we need the following observation: \par
For any $i \in \mathcal{K}$,
\begin{equation}
\label{alpha bound}
p_{i,t} \leq \sigma_t+   (k-K \sigma_t)\frac{(1-\sigma_t) \alpha_t }{\sum_{j\in \mathcal{K}} w_{j,t}^{\prime} }
\end{equation}
Now we let $\alpha_t=\min_{i \in \mathcal{K}} \left\{  w_{i,t}/(1-\sigma_t) \right\}$. Then we know that $\sum_{j\in \mathcal{K}} w_{j,t}^{\prime}=K \alpha_t (1-\sigma_t) $. It follows that the Right Hand Size (R.H.S) of the above inequality is equivalent to $\sigma_t +\frac{k-K \sigma_t}{K } $, by which we can ensure that $p_{i,t} \leq 1$ since $ k \leq K $.  As such, we conclude that there exist at least one $\alpha_t$ that ensures $p_{i,t} \leq 1$ for any $i$, which completes the proof.
\end{proof}
By Claim \ref{existence of inter alpha_t}, we justify the existence of $\alpha_t$ such that $p_{i,t} \leq 1$ holds for all $i$. But to determine our selection of $\alpha_t$, we should note that $\alpha_t$ should be set as large as possible \footnote{Too see why we make this statement, considering the case when setting $\alpha_{t} \to 0$, then the allocation differentiation is completely eliminated (i.e., all the clients share the same probability), which contradicts our tenet to give more chances to the stable contributors.}. As such, following the observation as per 
inequality (\ref{alpha bound}),  we indeed need to set $\alpha_{t}$ as follows :
\begin{equation}
\sigma_t+   (k-K \sigma_t)\frac{(1-\sigma_t) \alpha_t }{\sum_{j\in \mathcal{K}} w_{j,t}^{\prime} }=1
\end{equation}
By this equation, we maximize $ \alpha_t$ while making $p_{i,t} \leq 1 $ for any client $i$,  and it can be simplified as follows:
\begin{equation}
\label{aaaaaaee}
\frac{ \alpha_t }{\sum_{j\in \mathcal{K}} w_{j,t}^{\prime} }= \frac{1}{k- K \sigma_t}
\end{equation}
 Then we show the way to find the solution of Eq. (\ref{aaaaaaee}). First, it is noticeable that the possible “structures” of $\sum_{j\in \mathcal{K}} w_{j,t}^{\prime}$  are finite. Then we can simply divide it into at most $N-1$ cases. Let $\Psi_{i,t}= w_{i,t}/(1- \sigma_t)$. Formally, we assume \textit{case} $v$ satisfying: $\Psi_{i_v,t} \leq \alpha_{t}< \Psi_{i_{v+1},t}$ where $i_v$ denotes the $v$-th smallest $\Psi_{i,t}$. As the structure of  $\sum_{j \in \mathcal{K}} w_{j,t}^{\prime}$ is fixed for case $v$, we can derive:
\begin{equation}
\sum_{ w_{j,t} \leq \Psi_{i_v,t}} \frac{w_{j,t}}{ \alpha_{t} }  + \sum_{  w_{j,t} > \Psi_{i_v,t}} (1- \sigma_t)= k-K \sigma_t
\end{equation}
Reorganizing the term, we have:
\begin{equation}
\alpha_{t}= \frac{  \sum_{ j \in \mathcal{K}: w_{j,t} \leq \Psi_{i_v,t}} w_{j,t}   }{k-K\sigma_t- \sum_{  w_{j,t} > \Psi_{i_v,t}}(1 -\sigma_t) }
\end{equation}
By iterating all the cases and checking if the calculated $ \alpha_{t}$ satisfies the premise, i.e., $\Psi_{i_v,t} \leq \alpha_{t}< \Psi_{i_{v+1},t}$ , finally we are allowed to derive a feasible $ \alpha_{t}$.

\begin{algorithm}[h]  
        \caption {Exp3-based Client Selection (E3CS) for Federated Learning}
        \begin{algorithmic}[1] %每行显示行号  
        \REQUIRE ~~\\
        The number of involved clients each round; $k$ \\
         Fairness quota; $\{\sigma_t\}$\\
        Final round; $T$\\
        Local data distribution; $\{\mathcal{D}_i\}$\\
        Local update operation; $o_1(\cdot)$\\
  		
   	 \ENSURE~~\\
        Global network weights; $\bm \Theta_{T+1}$\\
       \STATE Initialize $w_{i,1}=1$ for $i=1,2,\dots, K$
   	\FOR{$t=1,2,\dots,T$}
		\STATE    $\bm p_t, S_t=\text{ProbAlloc}(k,\sigma_t,\{w_{i,t}\})$
		 \STATE  $A_t \sim \text{multinomialNR}( \bm p_t/k, k )$
		 {\color{black}
		 \FOR{client $i \in A_t$ \textbf{in parallel} }
		 \STATE $\bm \Theta_{i,t} = o_1 ( \bm \Theta_{t}, \mathcal{D}_i   )$
		\ENDFOR 
		 \STATE \textbf{for} client $i \in [K] $, $x_{i,t}=1$ if succeed; else, $x_{i,t}=0$ 
		\STATE $ \bm \Theta_{t+1}=\sum_{i\in A_t\text{and}x_{i,t} \neq 0} \omega_i \bm  \Theta_{i,t} +\sum_{i \notin A_t\text{or}x_{i,t} = 0} \omega_i  \bm \Theta_{t}$
		 \STATE $\hat{x}_{i,t}=\frac{\mathbb{I}_{ \{ i \in A_t\} }}{p_{i,t}}x_{i,t}$ \quad $i \in \mathcal{K}$}
		 \STATE  for $i \in \mathcal{K}$:  \begin{equation} \nonumber
w_{i,t+1}= \begin{cases} w_{i,t} \exp\left(\frac{(k-K\sigma_{t})\eta \hat{x}_{i,t}  }{K} \right )  &  i \notin S_t \\ w_{i,t} &  i \in S_t   \end{cases}
 \end{equation}
       \ENDFOR
        \end{algorithmic}
\label{E3CS}		
	\end{algorithm}

%\begin{algorithm}[h]  
%	
%	\caption {Local Update (LocUpdate)}
%	 \begin{algorithmic}[1] %每行显示行号  
%	{ \color{black}
%	\REQUIRE ~~\\
%	Global weights at round $t$; $\Theta_t$ \\
%	Local data set of client $i$; $\mathcal{D}_i$ \\
%	Local update epochs; $E_i$\\
%	mini-batch size; $B$ \\
%	Learning rate of optimizer; $\tilde{\eta}$
%	
%	 \ENSURE~~\\
%	Weights after local update; $\Theta_{i,t}$\\
%	\FOR{ epoch=$1,\dots,E_i$ }
%	\FOR{ each minibatch $B_c$ of local data $\mathcal{D}_i$ }
%	
%	\STATE {\color{red}$\Theta_{i,t}=\Theta_{t}-\frac{\tilde{\eta}}{B}\nabla_{\Theta_{i,t}} \mathcal{L}(\Theta_{i,t},\mathcal{B}_c)$}
%	 \IF{aggregation deadline is met}
%	\STATE Return $x_{i,t}=0$, $\Theta_{i,t}=\Theta_{t}$
%	\ENDIF
%	\ENDFOR
%	\ENDFOR
%	
%	\STATE Return $x_{i,t}=1$, $\Theta_{i,t}=\Theta_{i,t}$
%}
%\end{algorithmic}
%{\color{black}
%\textbf{Note:} more optimization techniques (e.g. momentum, variance reduction) could be potentially applied in the {\color{red}red line}.}
%	\label{local update}
%\end{algorithm}

\subsection{Algorithm} 
To enable a better understanding of our proposed selection solution, we present the detailed procedure of our Exp3-based Client Selection (E3CS) for Federated Learning in Algorithm \ref{E3CS}. Explicitly, E3CS runs in the following procedure:
\par
 \textbf{ Initialization.} Initialize the exponential weights of all the clients to 1 in the first round and then the algorithm formally goes into the iterative training process.
\par
 \textbf{ Probability allocation.}
In each iteration, Algorithm \ref{Probability allcoation alg}, which calculates the probability allocation as per we describe in section \ref{probability allocation section}, is called into execution and return the probability allocation $\{p_{i,t}\}$ and overflowed client set $S_t$.
\begin{algorithm}[h]  
	\caption {\color{black}Probability Allocation (ProbAlloc)}
	\begin{algorithmic}[1] %每行显示行号  
		\REQUIRE ~~\\
		The number of involved clients each round; $k$ \\
		Fairness quota; $\{ \sigma_t \} $\\
		Exponential Weight for round $t$: $\{w_{i,t}\}$
		\ENSURE~~\\
		Probability allocation vector for round $t$: $\bm p_{t}$\\
		Overflowed set for round $t$: $ S_{t}$\\
		\IF{  $\sigma_t+ (k- K\sigma_t)\frac{ \max_{i \in \mathcal{K}} \{w_{i,t} \} }{\sum_{j \in \mathcal{K}} w_{j,t} }>1$}
		
		\STATE Decide $\alpha_t$ such that: $\frac{ \alpha_t }{\sum_{j\in \mathcal{K}} w_{j,t}^{\prime} }= \frac{1}{k- K \sigma_t}$
		\STATE $S_t= \left \{ i \in \mathcal{K}: w_{i,t} >  (1-\sigma_t) \alpha_t  \right \}$
		\STATE $w_{i,t}^{\prime}=\min \{ w_{i,t}, (1-\sigma_t) \alpha_t\}$ for $i=1,2,...,K$
		\STATE $p_{i,t}= \sigma_t+ (k- K\sigma_t) \frac{w_{i,t}^{\prime}}{\sum_{j\in \mathcal{K}} w_{j,t}^{\prime} }$ for $i=1,2,...,K$
		\ELSE
		\STATE $S_t= \emptyset$ 
		\STATE $p_{i,t}= \sigma_t+ (k- K\sigma_t) \frac{w_{i,t}}{\sum_{j\in \mathcal{K}} w_{j,t} }$ for $i=1,2,...,K$
		\ENDIF 
		\STATE  Return $\bm p_{t}$, $S_t$
	\end{algorithmic} 	
	\label{Probability allcoation alg}
\end{algorithm}
\par
\textbf{ Stochastic selection.} {\color{black} Then our incoming task is to randomly select the clients based on the obtained probability. This specific task is done by drawing samples from a weighted multinomial distribution with no replacement for $k$ times. In our implementation, we simply use the API in PyTorch\footnote{\color{black} Similar implementation of multinomial sampling is also available in scipy, numpy and R.}, i.e., torch.multinomial($\bm p_t$, $k$, replacement=False)}. 
\par
 \textbf{ Local training and aggregation.}
{\color{black}Once $A_t$ is drawn from the multinomial distribution, our main algorithm would distribute the global model $\bm \Theta_t$ to the selected clients, who will immediately start local training (e.g., conduct multiple steps of SGD). This local training process could be different by using different optimizers (or adding a proximal term to the local loss function as FedProx \cite{li2018federated} does). At the same time, the server's main process sleeps until the deadline is met, and after that, models would be aggregated based on the returned models' parameters. Specifically, clients with successful returns would be involved in aggregation, and for those clients whose models are not successfully returned on time, or simply not being chosen into training, the global model would take its place in aggregation (see line 9 in Algorithm \ref{E3CS}). }
\par
\textbf{Exponential weights update.} After the aggregation is done, we need to update our "expectation" of clients (or arms). This part of update process is consistent with Eqs. (\ref{unbised estimation final}) and (\ref{weight update final}).

\subsection{Theoretical Regret Guarantee}
In this sub-section, we might need to evaluate our proposed solutions at a theoretical level. Let us first define an optimal solution of $P3$.
{\color{black}
\begin{definition}[Optimal solution] 
\label{optimal solution}
The optimal solution for P3 performs client selection based on the following probability allocation:
\begin{equation}
p_{i,t}^*= q_{i,t}^* \left (k-K\sigma_t \right)+ \sigma_t 
\end{equation}
where $q_{i,t}^*$ is the optimal allocation quota of $k-K\sigma_t$ probability.
\end{definition}
}
The definition is quite intuitive. We simply reserve  $ \sigma_t$  probability for each client and optimally allocate the residual $k-K\sigma_t$ probability. Based on our definition, it is not hard to derive the optimal  expected Cumulative Effective Participation until round $T$ ($\text{CEP}^*_T$):
\begin{equation}
\label{Ecep*}
\mathbb{E} [\text{CEP}^*_T]= \sum_{t =1}^T   \sum_{i \in \mathcal{K}} \left (q_{i,t}^* \left (k-K\sigma_t \right)+ \sigma_t \right) x_{i,t} 
\end{equation}
Given $\mathbb{E} [\text{CEP}^*_T]$, we like to compare the performance of E3CS with the optimal solution, which motivates us to give the following definition of regret:
\begin{definition}[Regret of E3CS]Given $\mathbb{E} [\text{CEP}^*_T]$, the regret of E3CS (or performance gap to the optimal) is given by 
\begin{equation}
\label{regret of E3CS}
R_{T}=\mathbb{E} [\text{\rm CEP}^*_T]- \mathbb{E} [ \text{\rm CEP}^{\rm E3CS}_T ]
\end{equation}
where $ \mathbb{E} [ \text{\rm CEP}^{\rm E3CS}_T ]=  \sum_{t =1}^T   \sum_{i=1}^K p_{i,t} x_{i,t} $.
\end{definition}
Now we introduce an upper-bound of the defined regret, as in Theorem \ref{regret bound theorem}.
\begin{theorem}[Upper bound of regret]
\label{regret bound theorem}
The regret of E3CS is upper-bounded by:
\begin{equation}
R_{T} \leq  \eta \sum_{t=1}^T   (k-K\sigma_t)  + \frac{K}{\eta} \ln{K}
\end{equation}
and if $\eta= \sqrt{\frac{K \ln K}{ \sum_{t=1}^T     (k-K\sigma_t)}}$,  we have:
\begin{equation}
R_{T} \leq   2\sqrt{ \sum_{t=1}^T K (k-K\sigma_t)      \ln K    } 
\end{equation}
\end{theorem}
\begin{proof} Complete proof is available in Appendix \ref{regret proof}. \end{proof}
\begin{remark}
Notably, the regret of E3CS would diminish to 0 as $\sigma_t \to k/K$. This phenomenon can be explained by looking into what happens if $\sigma_t = k/K$: both E3CS and the optimal solution yield the same solution, which is essentially an even random selection among clients, so the regret would be exactly 0. At the other extreme when $\sigma_t =0$ for all $t$, the regret would be reduced to $R_{T} \leq 2\sqrt{ T K k \ln K} $ if $\eta=\sqrt{\frac{K \ln K}{ Tk}}$. In this case, the derived bound of regret under multiple play setting is $\sqrt{k}$ times of that of the canonical single play Exp3 (see Theorem 11.1 in \cite{lattimore2020bandit})
\end{remark}
\section{ Experiments }
In this section, we shall present the experimental results of our proposed client selection solution. The evaluation is based on numerical simulation and real training in two iconic public datasets: EMNIST-Letter \cite{cohen2017emnist} and CIFAR-10\cite{krizhevsky2009learning}.
\subsection{Setup}
In this subsection, we shall illustrate the basic setup of our experiment.

 \textbf{Programming and running environment.}
We have implemented E3CS and the volatile training feature to an open-source light-weighted federated learning simulator named FlSim \footnote{Source code available in \url{https://github.com/iQua/flsim}}, which is built on the basis of PyTorch and has efficiently implemented parallel training for the learning process. In terms of the runtime environment, all the computation in our simulation is run by a high-performance workstation (Dell PowerEdge T630 with 2xGTX 1080Ti).

 \textbf{Simulation of volatile and heterogeneous clients.}
In our simulation, a total number of $K=100$ volatile clients are available for selection, and in each round, $k=20$ of which are being selected.
Under a volatile training context, clients might suffer unexpected drop-out during their training. In our experiment, we adopt a Bernoulli distribution to simulate the training status of clients. Formally, we have $x_{i,t} \sim \text{Bern}( \rho_i)$ where $\rho_i$ is the success rate of client $i$. {\color{black}To simulate the \emph{heterogeneous volatility} of clients, we equally divide the whole set of clients into 4 classes, with the success rate respectively set as 0.1, 0.3, 0.6, and 0.9. In addition, we also allow the local training epochs of clients to be different to simulate the client's \emph{heterogeneous computing capacity}. Specifically, we designate different training epochs for different clients, and the designated epochs are randomly chosen from the set \{1,2,3,4\}. But please note that we do not introduce correlation between \emph{heterogeneous volatility} and \emph{heterogeneous computing capacity}. That is, a client that is asked to perform more rounds of training does not necessarily lead to a higher or lower failure rate. These two properties are assumed to be independent in our analyzed scenario.}

 \textbf{Simulation of data distribution.}
We simulate the data distribution by both iid and non-iid setting\footnote{\color{black} In federated learning, each client only has a small number of data for training. To mimic the federated setting, we split the whole training dataset (like 50000 images in CIFAR-10) into small portions (like 500 images per client), and send each portion to different clients to simulate distributed training environment. Also, in federated learning, the data of clients are usually heterogeneous. For example, consider a simple MNIST FL task, in which the data of a client is mostly labeled as "1", while that of another is mostly labeled as "2". Then two of the clients are said to have non-iid data, and a similar data distribution like this is pretty common in reality (due to personalization). As such, we attempt to simulate federated learning in both iid (the ideal learning case) and non-iid setting (the more realistic case).}. 1) For an iid one, each client independently samples $|\mathcal{D}_i|$ pieces of data from the data set. 2) For a non-iid distribution, we randomly select one primary label for each client. Then we sample $0.8 \times |\mathcal{D}_i|$ pieces of data from those that coincide with their primary label and $0.2 \times |\mathcal{D}_i|$ from those with the remaining labels. {\color{black}For both the iid and non-iid settings, each client randomly reserves 10\% of the data for testing.} 
\begin{table}
 \small
\caption{Parameters Setting of FL training. Please note 1) Symbols that do not appear in other places of this paper would be written as "-". 2)SGD is abbreviation of standard Stochastic Gradient Descent. }
\label{Parameters setting}
\centering
\begin{tabular}{p{2.5cm}lll}
\hline
 Setting  & Symbols & Task 1 & Task 2\\ 
\hline
  Dataset   &-&EMNIST-Letter  &CIFAR-10\\
\# of labels &-&26 &10\\
 Max \# of rounds &$T$& $400$ & $2500$\\
\# of data per client & $|D_i|$ &$500$ & $500$\\
 Learning rate of E3CS & $\eta$ &$0.5$ & 0.5\\
 Optimizer& - &SGD & SGD \\
Learning rate of optimizer & -&$1e{-2}$ & $1e{-2}$\\
 Momentum &-& 0.9  & 0.9\\
 Local epoches &$E_i$& \{1,2,3,4\}  & \{1,2,3,4\}\\
 Mini-batch size & $B$ & 40  & 40\\
\hline
\end{tabular}
\end{table}

\textbf{Datasets and network structure.}
To evaluate the training performance, we prepare two tasks for FL to conquer: EMNIST-Letter and CIFAR-10. For EMNIST-Letter, we employ a CNN with two 5x5 convolution layers (each with 10 channels), followed by 2x2 max pooling and two fully connected layers with respective 1280 and 256 units, and finally a softmax output layer. For CIFAR-10, which is known to be a harder task, we use another CNN model with two 5x5 convolution layers (each with 64 channels), also followed with 2x2 max pooling, two fully connected layers with 384 and 192 units, and finally a softmax output layer. Other training-related setting is available in Table \ref{Parameters setting}.

\textbf{Baselines and implementation details.}
We prepare three state-of-the-art client selection scheme, namely FedCS\cite{nishio2019client}, Random\cite{mcmahan2016communication}, and pow-d \cite{cho2020client} for baselines. Details description of these baselines is available in Appdendix \ref{baseline description}.
Based on different settings of $\sigma_t$, we like to evaluate the performance of the following selection schemes:
\begin{itemize}
\item \textbf{E3CS- (number).} fairness quota $\sigma_t$ of E3CS is stationary and fixed to $\text{(number)} \times k/K$. In this setting, the fairness constraint would be ineffective and the algorithm would seek to maximize the eCEP regardless of fairness.
\item \textbf{E3CS-inc.} we allow fairness quota $\sigma_t$ to be incremental with training round $t$. More explicitly, we let {\color{black}$\sigma_t=k/K$} if $T \geq t>T/4$ and $\sigma_t=0$ if $1\leq t \leq T/4$. The motivation of our setting is that we want the algorithm to gain more effective participation in the beginning stage, but when training is approaching convergence, it should be better to expand the selection scope (i.e., to be fair) in order to improve the final accuracy. We will formally discuss the motivation behind this setting after we present the training result.
\end{itemize}
{\color{black}
Besides, as we note that we in this paper only focus on the client selection sub-problem, we need to borrow the update schemes (see $o_1$ in our global problem \textit{P1}) from the state-of-the-art methods. Explicitly, we mimic these two schemes:
\begin{itemize}
	\item \textbf{FedAvg\cite{mcmahan2016communication}.}  The local update scheme in FedAvg is SGD with cross entropy loss function.
	\item \textbf{FedProx\cite{li2018federated}.} The local update scheme in FedProx is also SGD, but with a regularization invovled loss function. It adds a proximal term $\frac{\gamma}{2} ||\bm \Theta_{i,t}- \bm \Theta_{t}||^2$ into the typical cross entropy loss. 
\end{itemize}
In our simulation. we use "scheme+(A)" to denote the selection scheme with update operation  in FedAvg (e.g., Random(A), E3CS-0(A), etc). Similarly, we use "scheme+(P)" to denote the scheme with the same operation with FedProx (e.g., Random(P), E3CS-0(P), etc). For all the FedProx-based schemes, the proximal coefficient is set to $\gamma=0.5$.
}  

\subsection{Numerical Simulation}
{\color{black}
In this subsection, we present the numerical evaluation to show how different selection algorithms perform in terms of  important statistics like effective participation, times of selection, and we further discuss the results combining the working patterns of these algorithms.  
}    
\begin{figure*}[!t]
\centering
\subfloat{\includegraphics[width=2.2in]{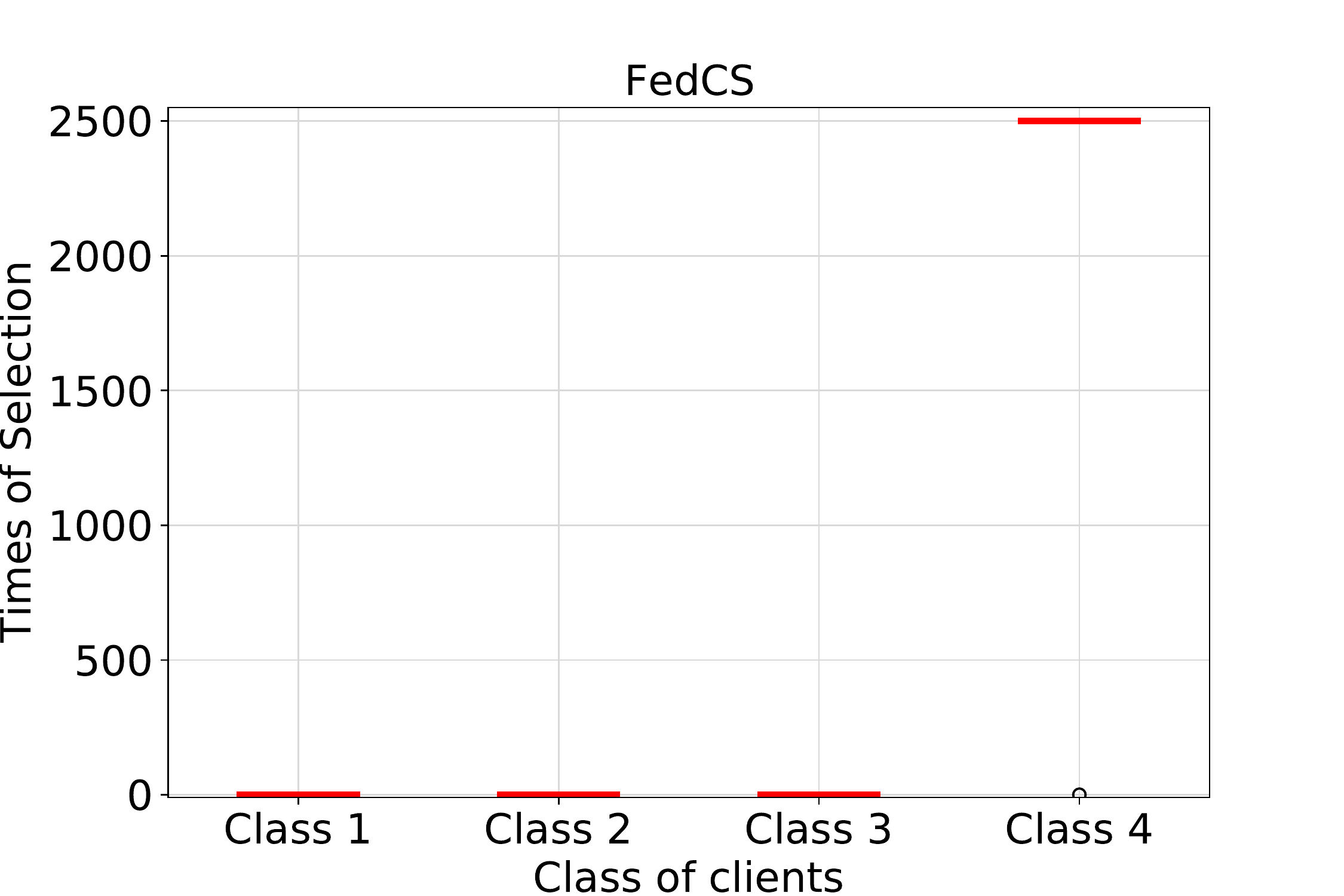}%
}
\subfloat{\includegraphics[width=2.2in]{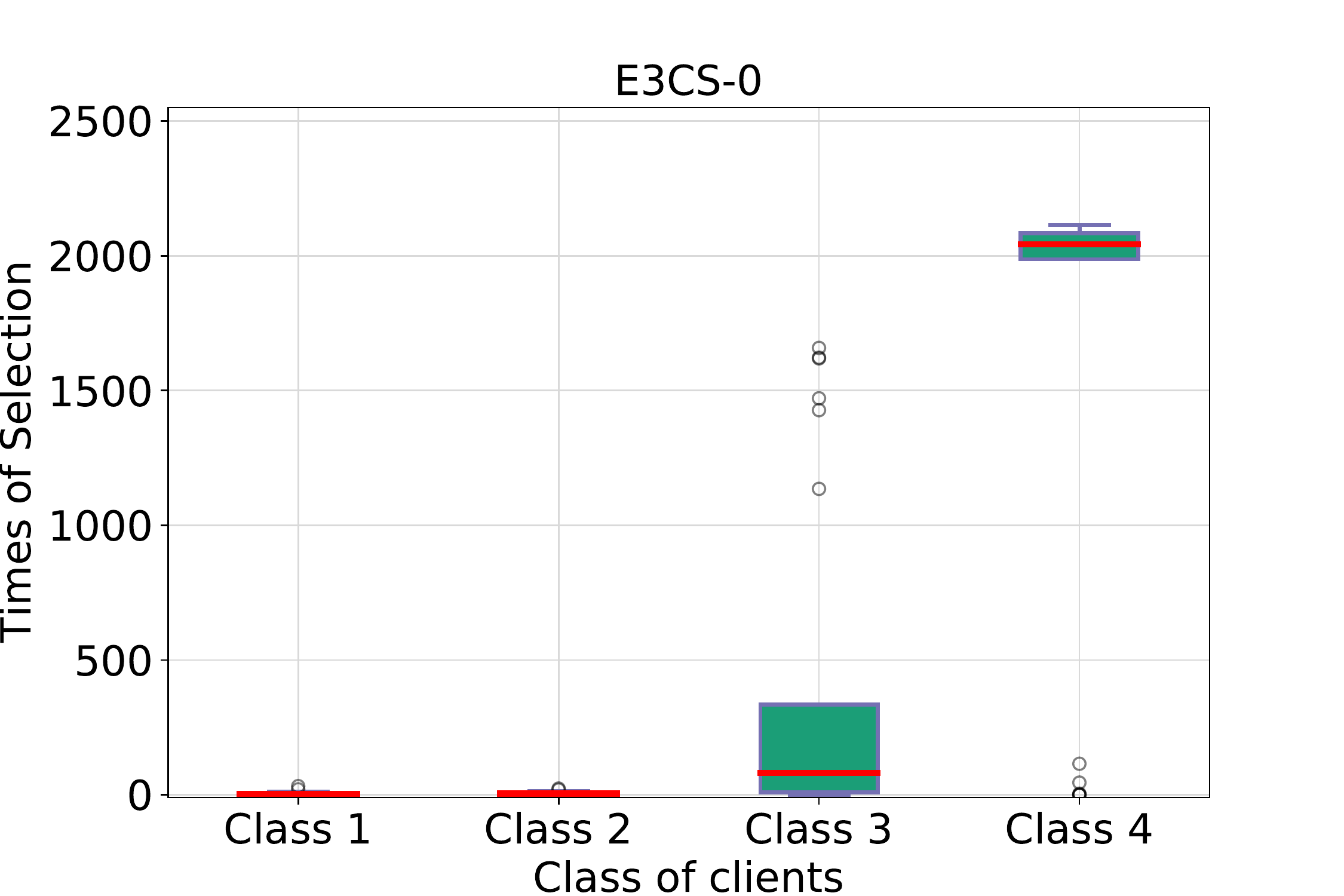}%
}
\subfloat{\includegraphics[width=2.2in]{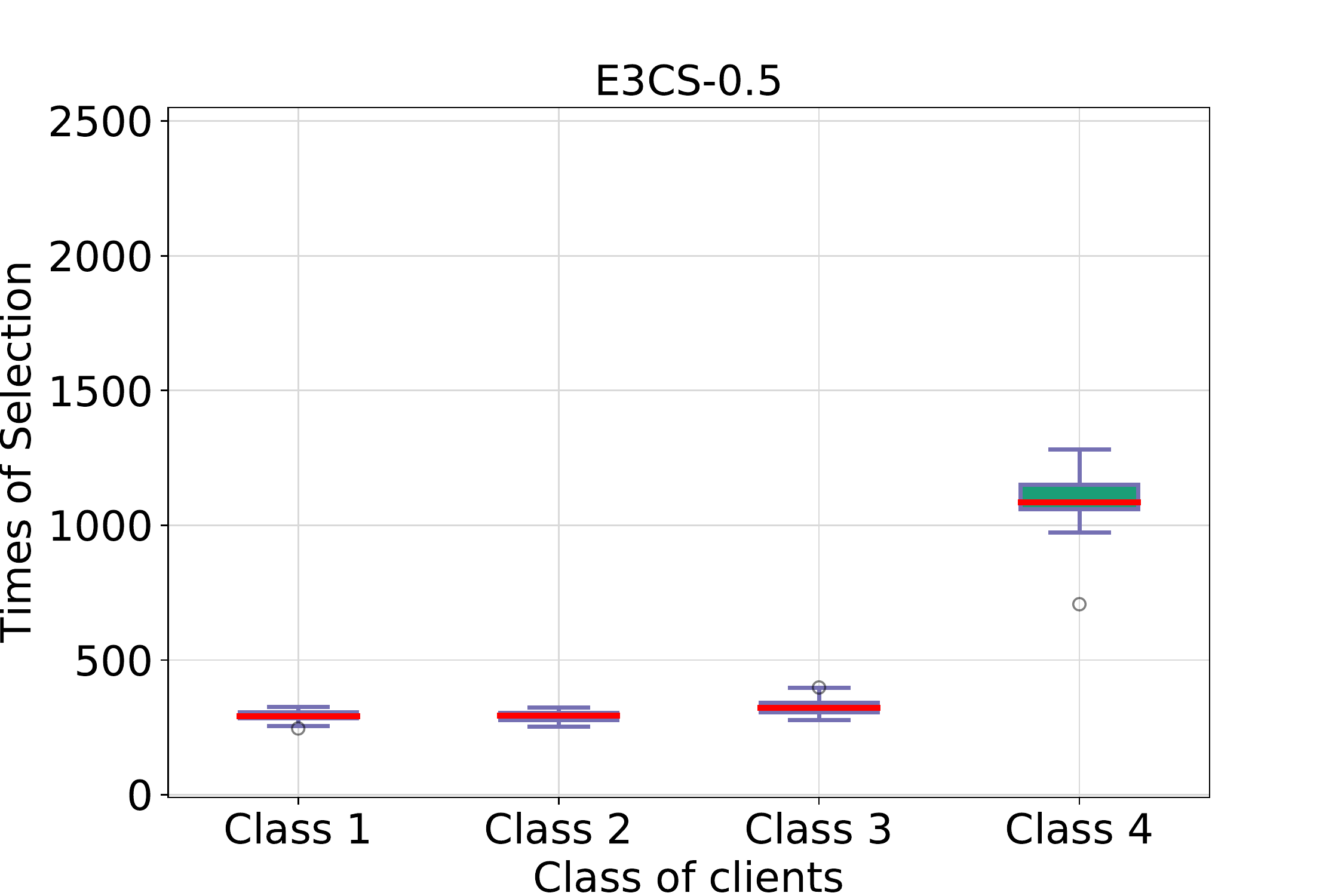}%
}\\
\subfloat{\includegraphics[width=2.2in]{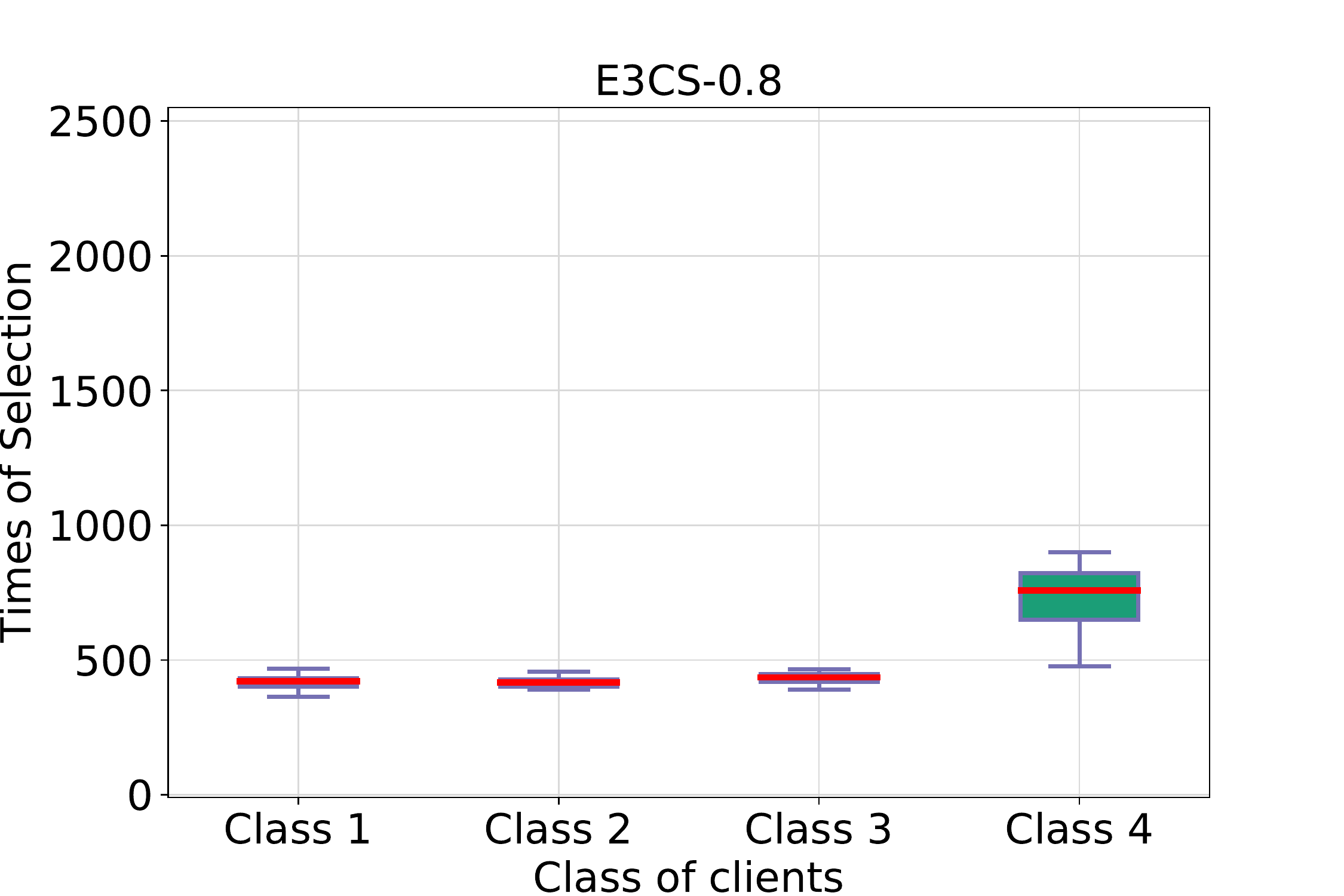}%
}
\subfloat{\includegraphics[width=2.2in]{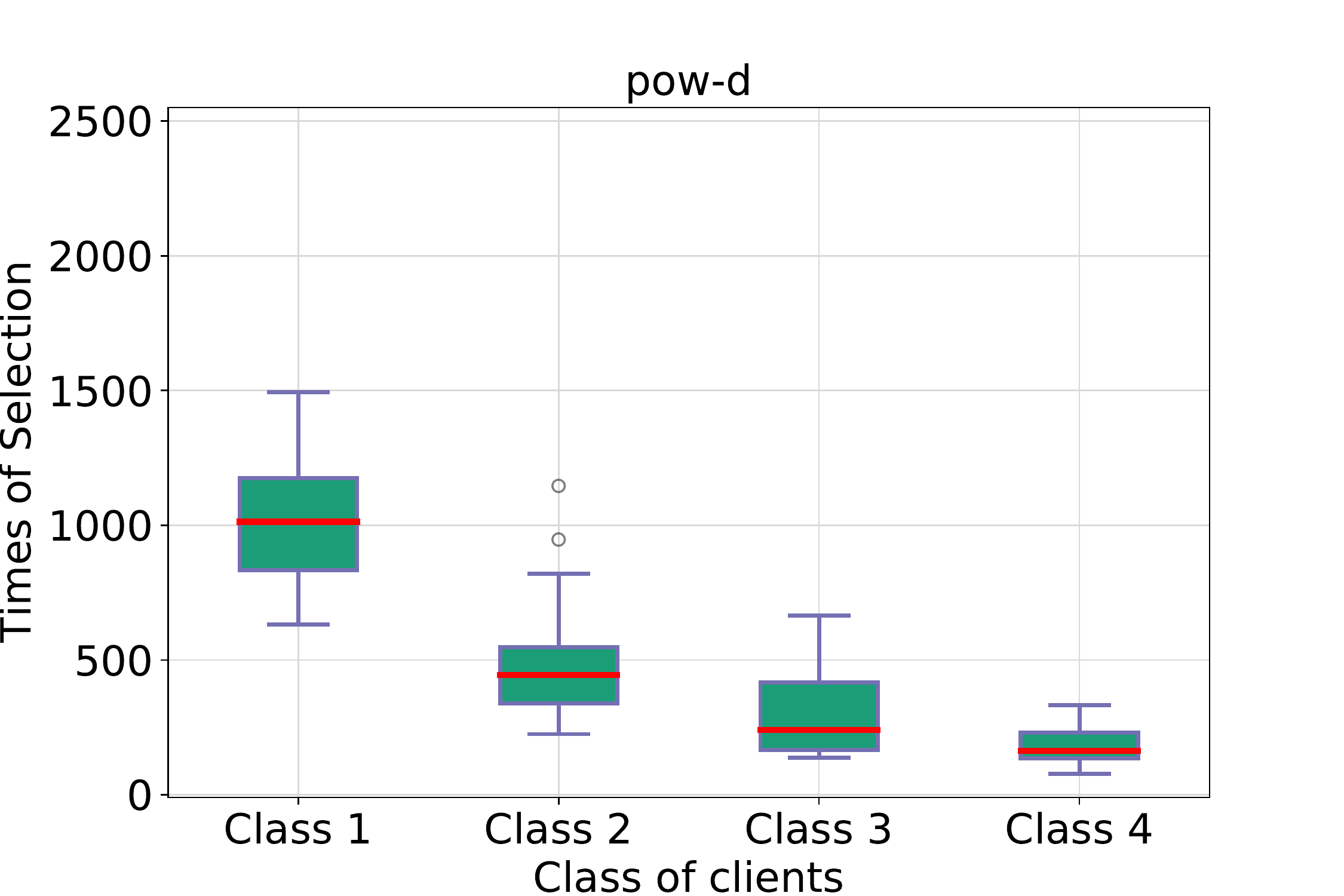}%
}
\subfloat{\includegraphics[width=2.2in]{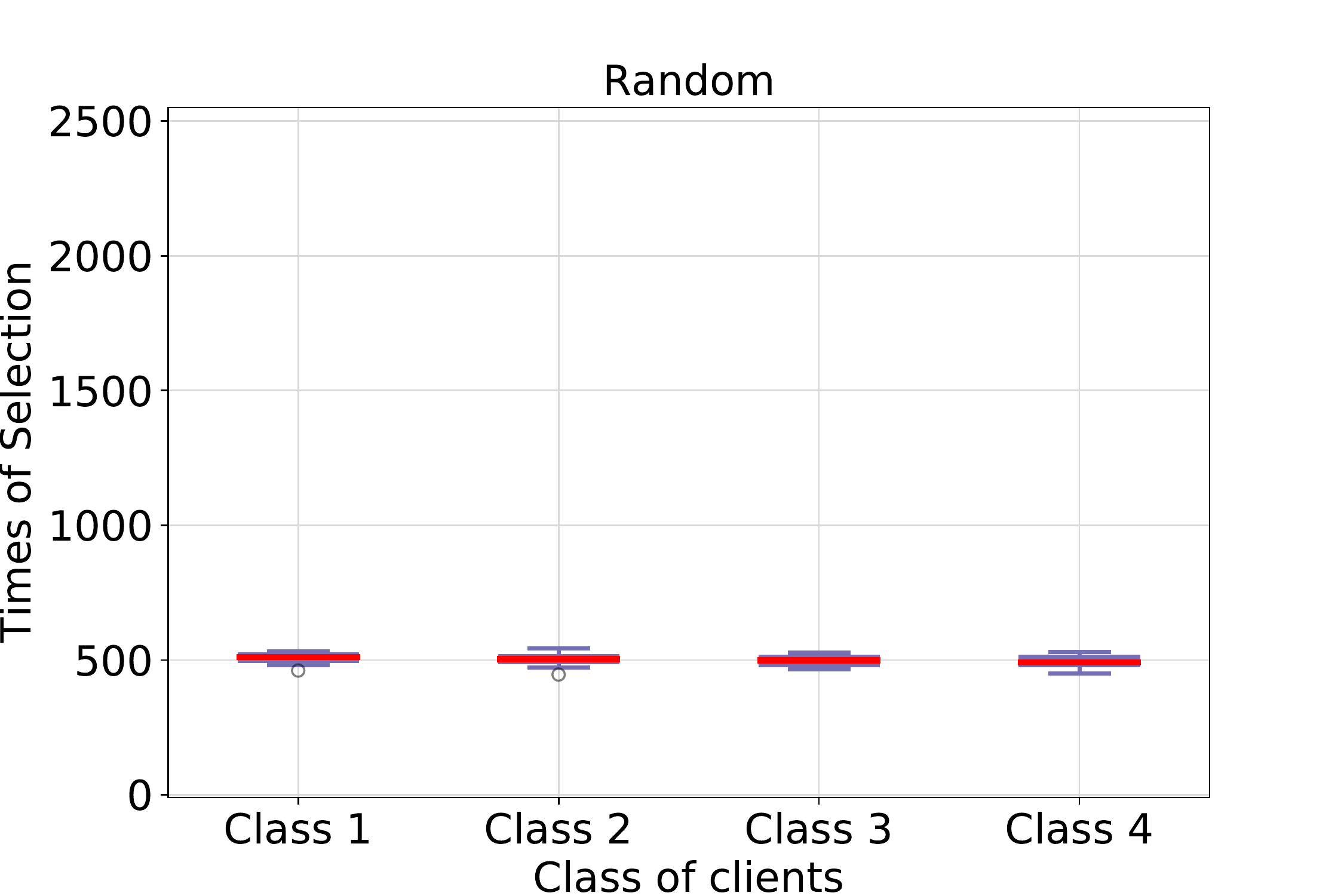}%
}
\caption{Under different selection schemes, the box plots display times of selection over 4 classes of clients, respectively with success ratio 0.1, 0.3, 0.6, 0.9, and an equal division of a total number of 100 clients. The total simulated communication rounds are all set to 2500 and in each round, 20 clients are to be selected. Please note that for FedCS scheme, all the selections are dedicated towards 20 of 25 {\color{black}class 4} clients, so the Inter-Quartile Range (IQR) of {\color{black}class 4} (which is not displayed in the plot) is 2500, while the IQR of other classes are 0.  }
\label{selection record}
\end{figure*}

 By running 2500 rounds of simulation step, we obtain the selection records of different selection schemes over different groups of clients. The results are formally displayed in several equal-scale box plots (see Fig. \ref{selection record}). From the figure, we can derive the following observations.
 
 {\color{black}\textbf{Effectiveness of E3CS in choosing reliable clients.} } One observation is that E3CS algorithms are able to learn the most reliable clients by only a modicum number of tries.  E3CS-0, a scheme that has no regard for fairness, only wrongly selects sub-optimal classes of clients dozens of times over a total of 2500 rounds of selection. Besides, by comparison of E3CS-0 and FedCS, we can observe another unexpected advantage of E3CS, i.e., a certain degree of fairness would be reserved if a sufficient amount of clients shares the same (or near) high success ratio. Take our setting as an example. In our setting, FedCS would consistently choose specific 20 out of 25 clients belonging to {\color{black}Class 4}, but by contrast, E3CS-0 would share the most probability over all the 25 clients in {\color{black}Class 4} while giving minor probability (resulting from the cost of learning) to others classes of clients. This selection mechanism gives some degree of fairness while not necessarily sacrificing a lot in terms of CEP. It can be observed that pow-d is prone to select the clients which are more likely to fail, which is directly against the pattern of E3CS and FedCS. This phenomenon can be explainable by the following analysis. 1) By its objective, we know that pow-d tends to select the clients with higher losses. 2) And those clients that are more likely to fail typically have a higher loss, since their local model has less chance to be aggregated into the global model. Consequently, clients with higher failure probability are more favored by pow-d.
 
\textbf{Selection fairness.} As depicted, all the schemes barring Random and pow-d would deliver more selection frequency to the clients in {\color{black}Class 4}, who boast the highest success rate. By this biased selection, Cumulative Effective Participation (CEP) could be maximized, but fairness degree might correspondingly hurt. Obviously, the order of fairness degree among all the selection schemes is: Random $> $ E3CS-0.8 $>$ {\color{black}pow-d}  $>$ E3CS-0.5 $ >$ E3CS-0 $>$ FedCS. This particular order of fairness might serve a critical function to reach effective training, which would be specified later.
\begin{figure}[!hbtp]
\centering
\includegraphics[width=3.5in]{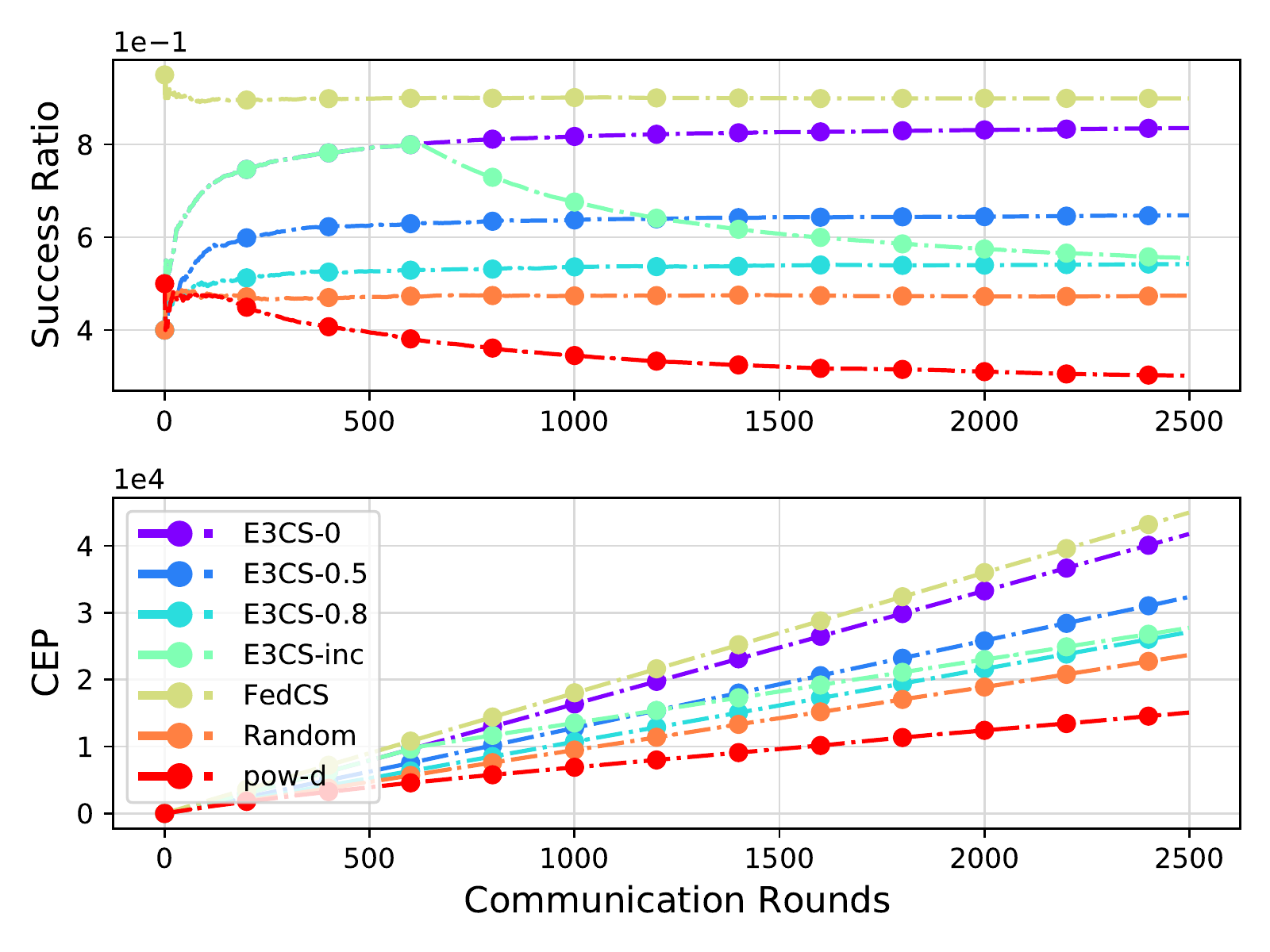}
% where an -eps-converted-to.pdf filename suffix will be assumed under latex, 
% and a .pdf suffix will be assumed for pdflatex; or what has been declared
% via \DeclareGraphicsExtensions.
\caption{Communication Rounds vs. Success Ratio and Cumulative Effective Participation (CEP)  for different selection schemes. }
\label{success ratio}
\end{figure}
\par

\textbf{Success ratio and CEP. } We define two metrics to evaluate the proposed solutions:  i) success ratio, explicitly formulated as $\sum_{t=1}^T \sum_{i \in A_t}x_{i,t}/ T k $ and ii) the Cumulative Effective Participation (CEP), formulated as, $\sum_{t=1}^T\sum_{i \in A_t}x_{i,t}$. We present the evolvement of these two quantities with communication rounds, as depicted in Fig. \ref{success ratio}. 
From the top sub-figure, it is notable that the success ratio of all the E3CS (except E3CS-inc) shares a similar trend of convergence and the final convergence value is largely determined by the constant setting of $\sigma_t$, indicating that fairness quota has a negative correlation on training success ratio. This phenomenon is well justified since a larger fairness quota might necessarily expand the selection of clients that are prone to fail! Another interesting observation is that  E3CS-inc undertakes a significant turn in Round 625 i.e., the exact round of $4/T$. This phenomenon is  as expected since E3CS-inc essentially reduces to an unbiased random selection scheme after $4/T$ and therefore its success ratio would be correspondingly plunged. Theoretically, E3CS-inc will eventually converge to the convergence value of Random if $T$ goes sufficiently large.  Also, it is interesting to find that pow-d has a very low CEP and success ratio throughout the whole training session. This experimental result is in accordance with our previous analysis, in that pow-d is prone to select clients with lower success rates.

\par 

\subsection{Real Training on Public Dataset}
We then show our experimental results of real training on public dataset EMNIST-Letter \cite{cohen2017emnist} and CIFAR-10\cite{krizhevsky2009learning}.
\begin{figure*}[!t]
	\centering
	\subfloat[\color{black} iid, FedAvg-based ]{\includegraphics[width=1.8in]{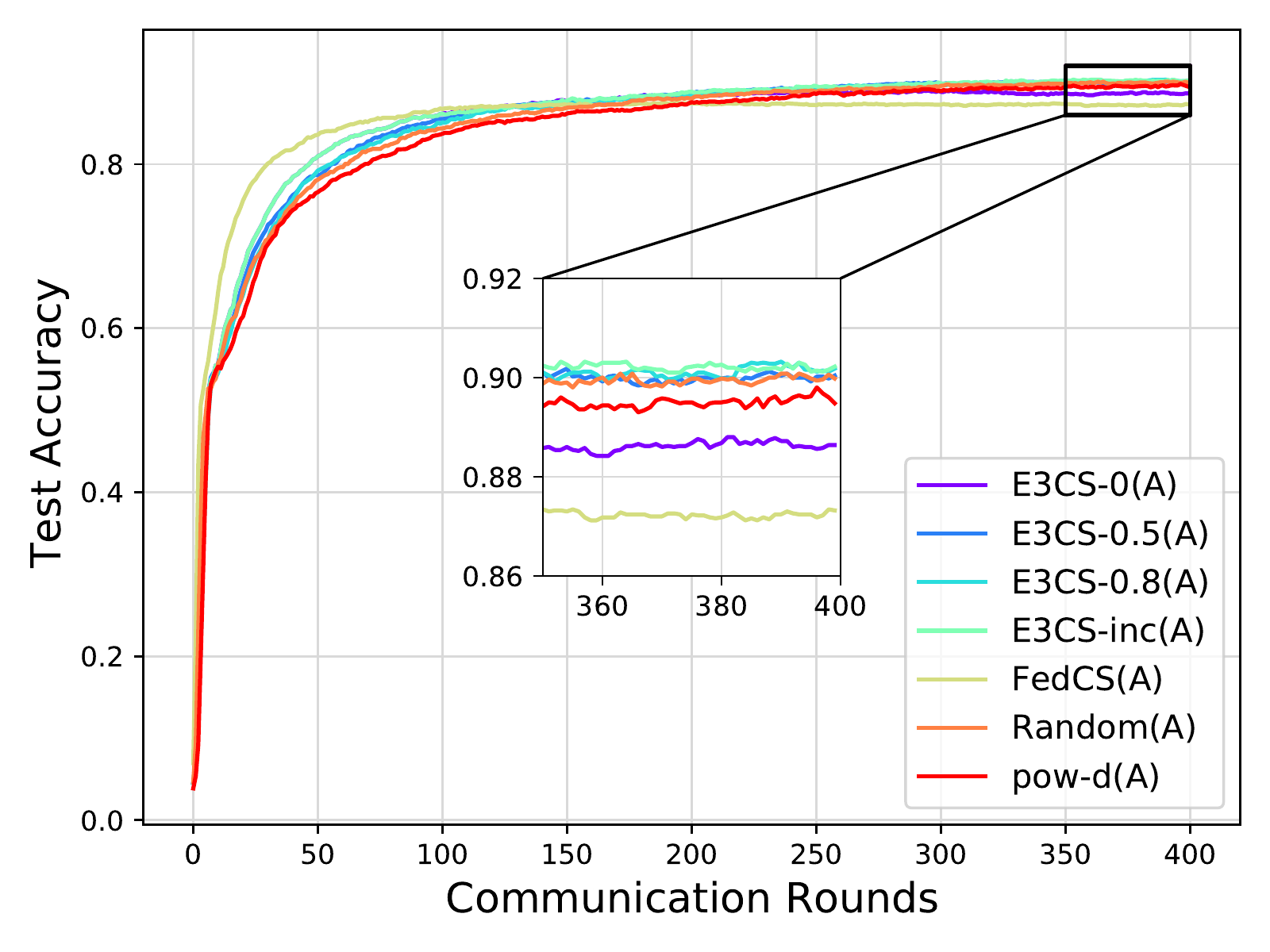}%
	}
	\subfloat[\color{black} non-iid, FedAvg-based ]{\includegraphics[width=1.8in]{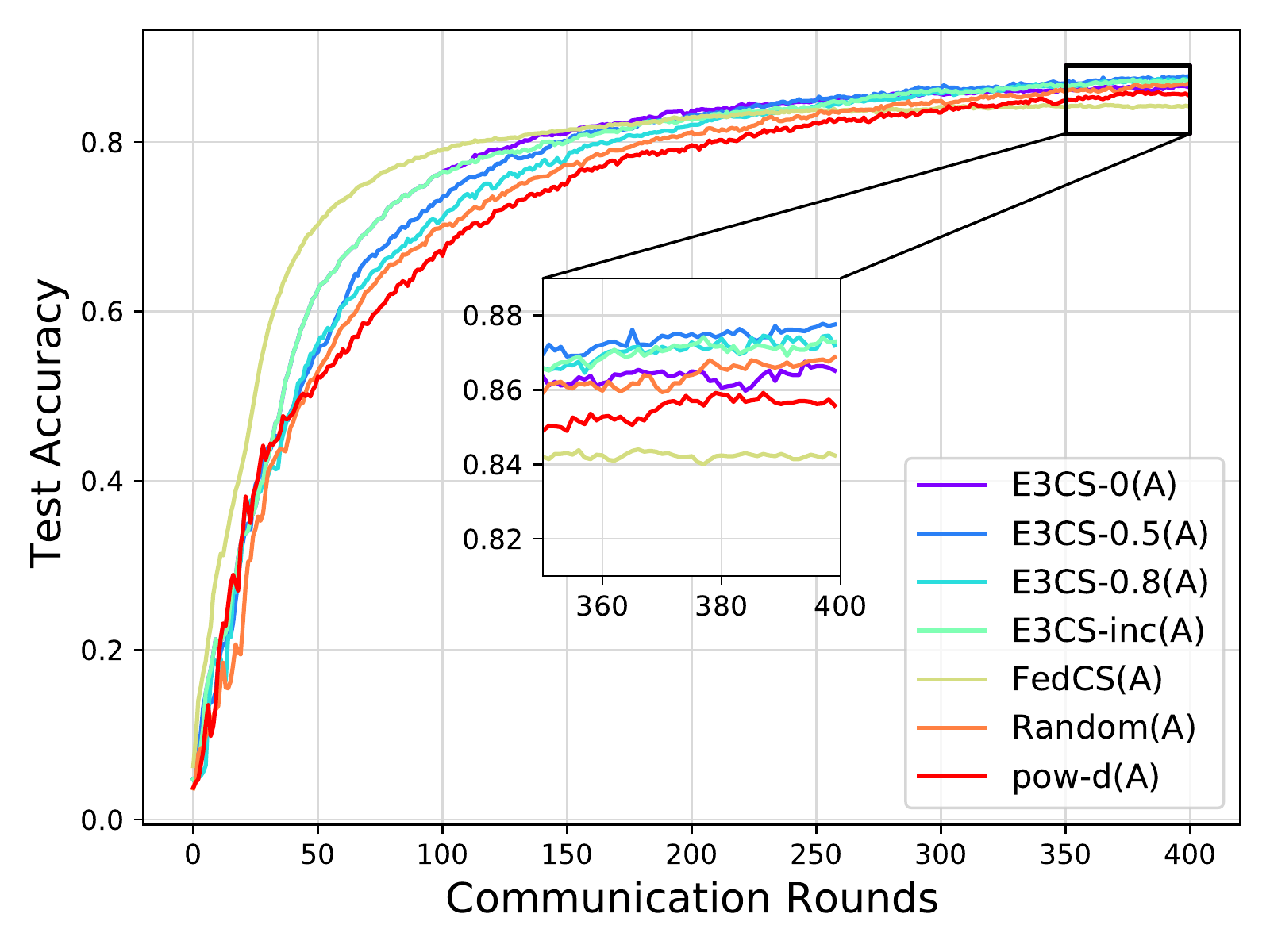}%
	}	
	\subfloat[\color{black} iid, FedProx-based ]{\includegraphics[width=1.8in]{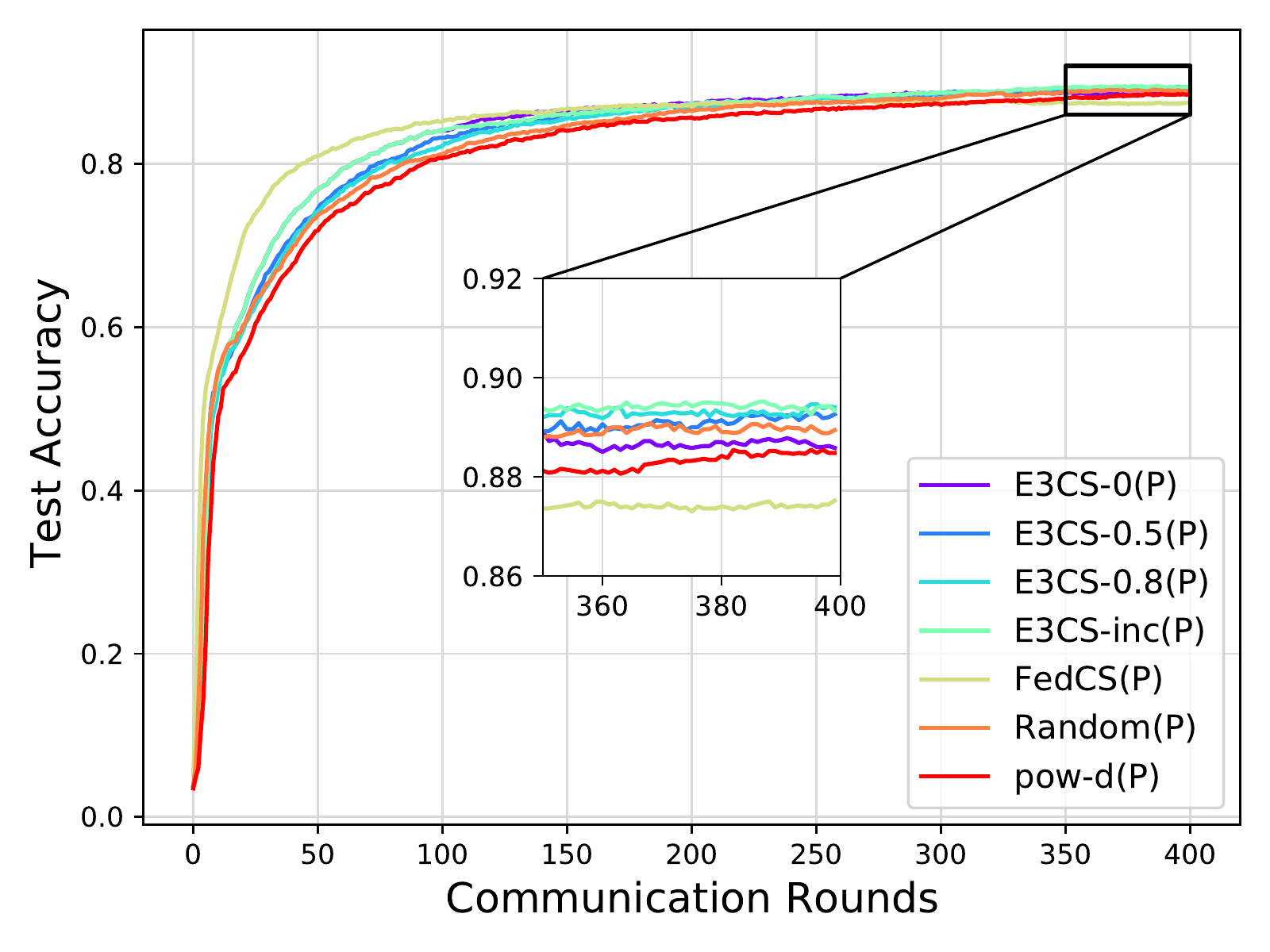}%
	}
	\subfloat[\color{black} non-iid, FedProx-based ]{\includegraphics[width=1.8in]{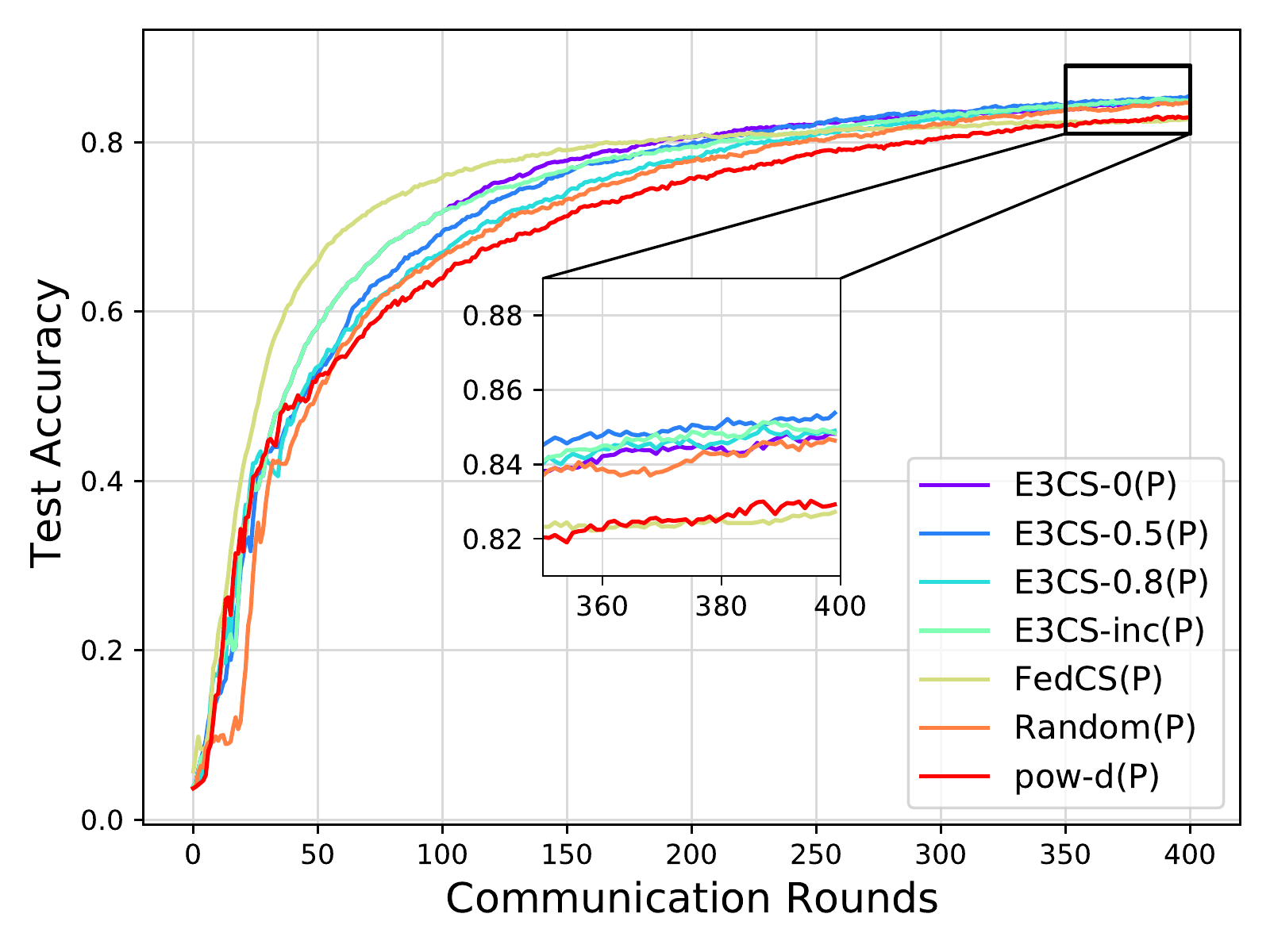}%
	}
	\caption{Test accuracy vs. communication rounds for EMNIST-Letter. }
	\label{acc EMNIST}
\end{figure*}
\begin{table*}[!t]
\caption{Performance evaluation for EMNIST-Letter. Please note: 1) Accuracy@\textbf{number} represents the first round to reach a certain test accuracy.   2) NaN means that the accuracy never reach the corresponding setting over the training session (400 rounds).  }
\label{ result EMNIST-Letter}
\centering
\color{black}{
\begin{tabular}{ccc|cc|cc|cc}
\midrule \multirow{2}{*}{Methods}     & \multicolumn{2}{c}{Accuracy@\textbf{65}}  & \multicolumn{2}{c}{Accuracy@\textbf{75}}  &  \multicolumn{2}{c}{Accuracy@\textbf{85}} &  \multicolumn{2}{c}{Final Accuracy (\%)}\\     \cmidrule(lr){2-3}    \cmidrule(lr){4-5}  \cmidrule(lr){6-7} \cmidrule(lr){8-9}         & iid    &   non-iid         & iid    &   non-iid          & iid    &  non-iid       & iid    &  non-iid  \\ 
\midrule
E3CS-0(A)& 18& 58& 32& 94& 82& 257& 88.64& 86.52\\
E3CS-0.5(A)& 20& 67& 38& 107& 94& \textbf{243}& 90.04& \textbf{87.76}\\
E3CS-0.8(A)& 21& 75& 39& 117& 98& 272& 90.18& 87.2\\
E3CS-inc(A)& 18& 58& 32& 94& 82& 268& \textbf{90.22}& 87.3\\
FedCS(A)& \textbf{11}& \textbf{39}& \textbf{20}& \textbf{69}& \textbf{66}& NaN& 87.32& 84.24\\
Random(A)& 21& 78& 40& 131& 107& 312& 89.98& 86.88\\
pow-d(A)& 24& 92& 42& 148& 119& 339& 89.48& 85.58\\
\midrule
E3CS-0(P)& 24& 69& 45& 120& 112& NaN& 88.58& 84.82\\
E3CS-0.5(P)& 27& 82& 52& 140& 131& \textbf{375}& 89.26& \textbf{85.38}\\
E3CS-0.8(P)& 30& 89& 53& 157& 137& 387& \textbf{89.4}& 84.9\\
E3CS-inc(P)& 24& 69& 45& 132& 127& 386& 89.34& 84.84\\
FedCS(P)& \textbf{15}& \textbf{48}& \textbf{28}& \textbf{93}& \textbf{93}& NaN& 87.52& 82.72\\
Random(P)& 29& 92& 57& 167& 157& NaN& 88.94& 84.64\\
pow-d(P)& 34& 104& 63& 192& 171& NaN& 88.48& 82.92\\
\midrule
\end{tabular}
}
\end{table*}

\begin{figure*}[!hbtp]
	\centering
	\subfloat[\color{black}iid, FedAvg ]{\includegraphics[width=1.8in]{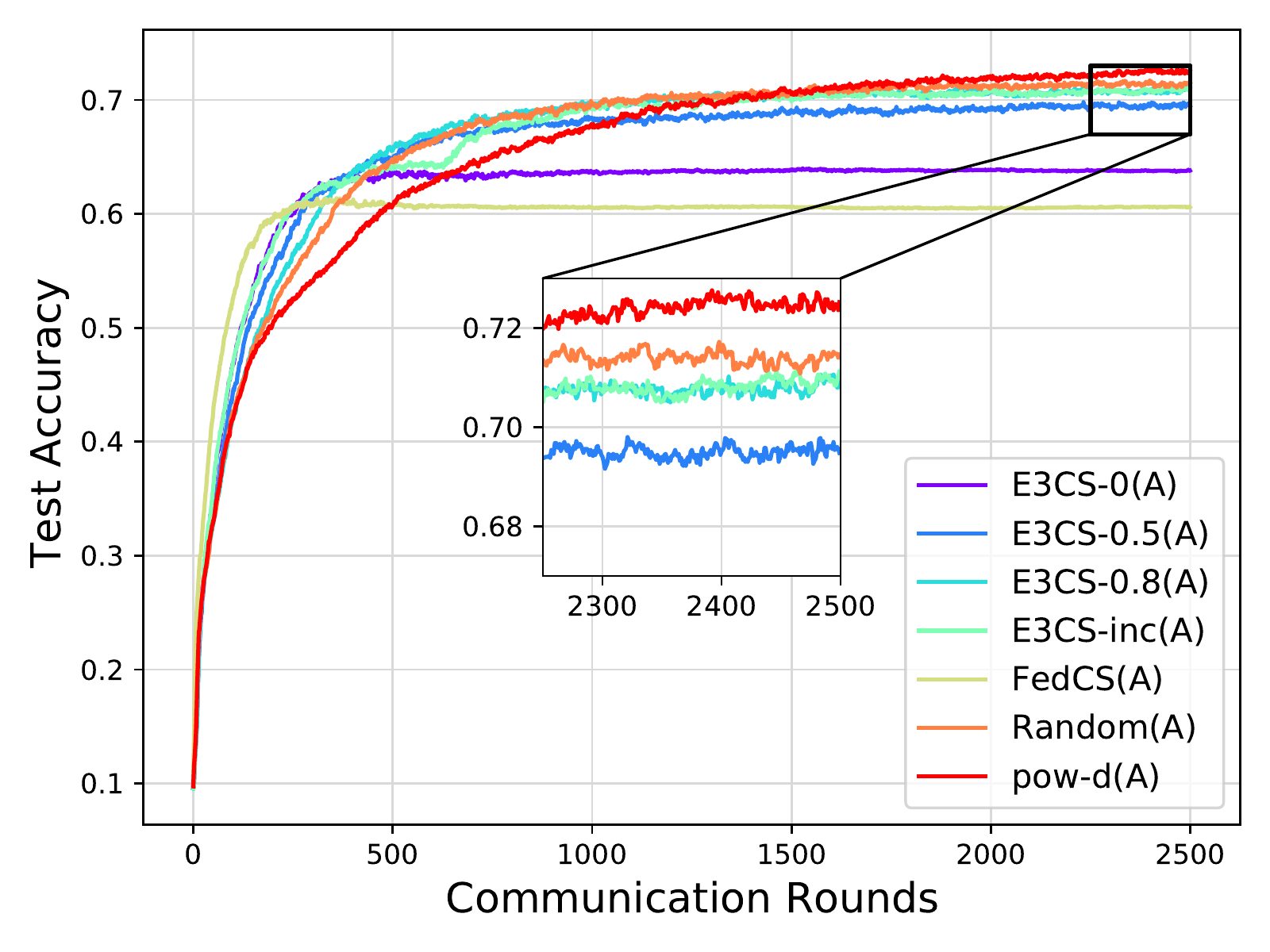}%
	}
	\subfloat[\color{black}non-iid, FedAvg ]{\includegraphics[width=1.8in]{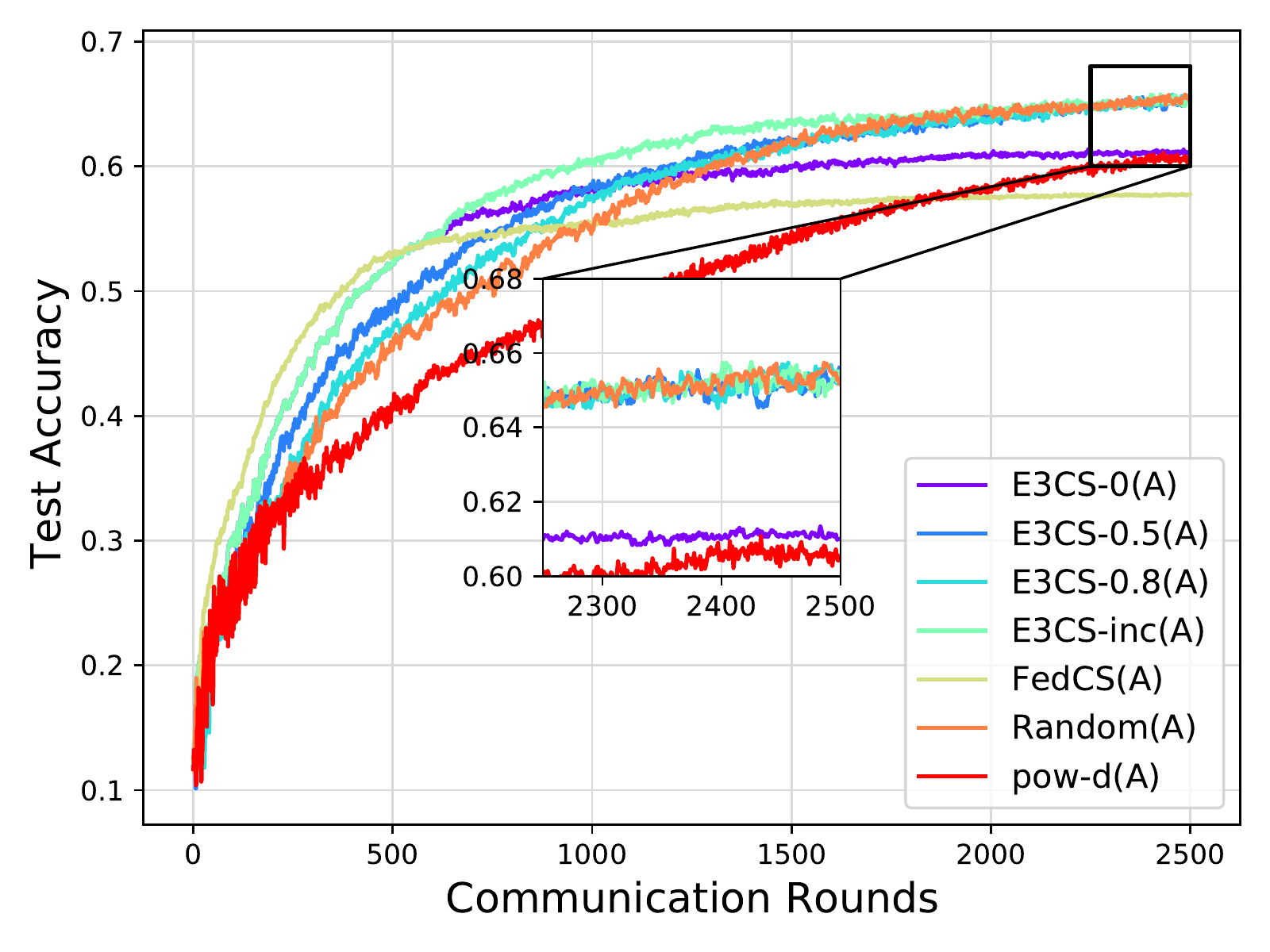}%
	}	
	\subfloat[\color{black} iid, FedProx ]{\includegraphics[width=1.8in]{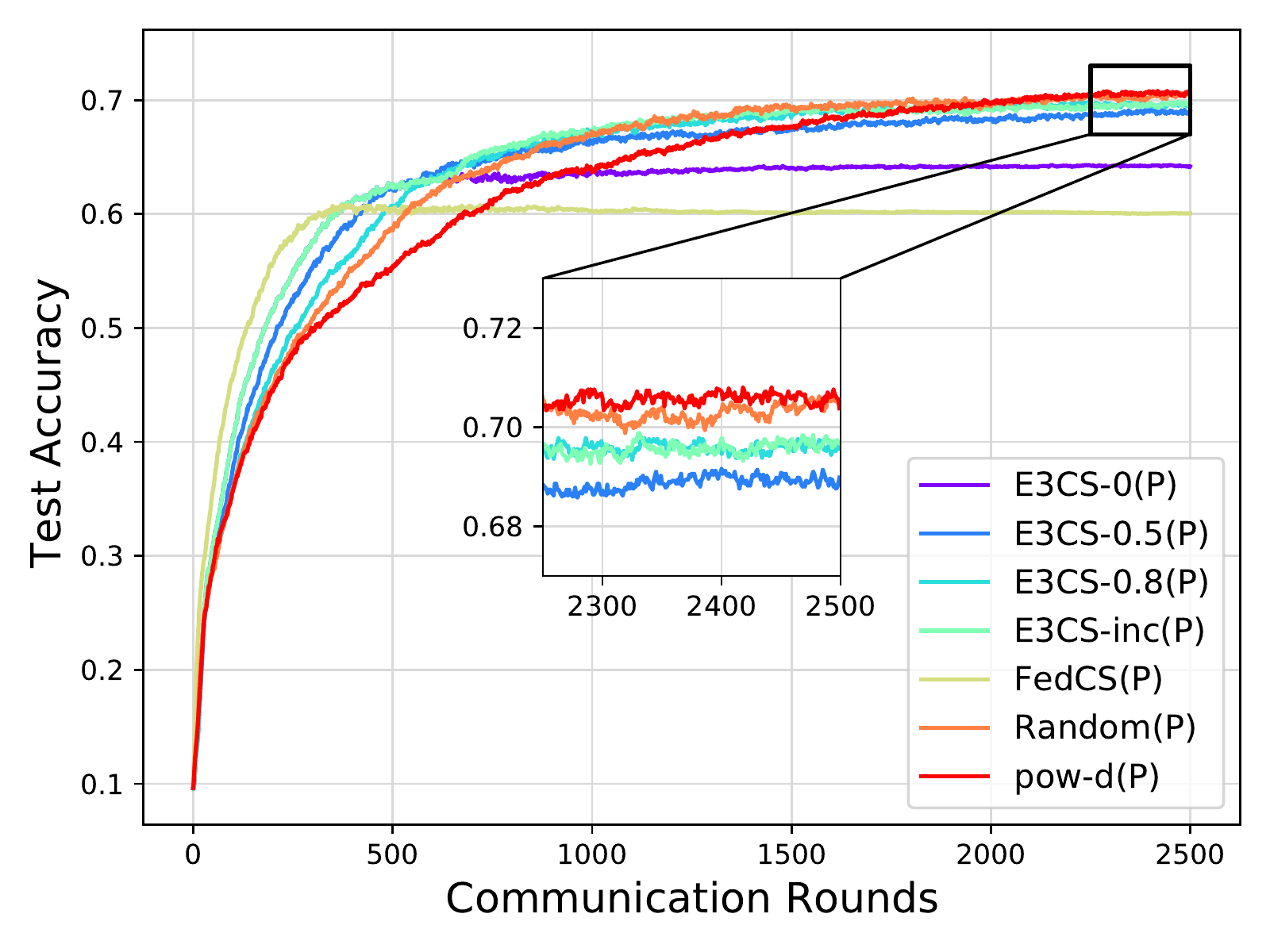}%
	}
	\subfloat[\color{black} non-iid, FedProx ]{\includegraphics[width=1.8in]{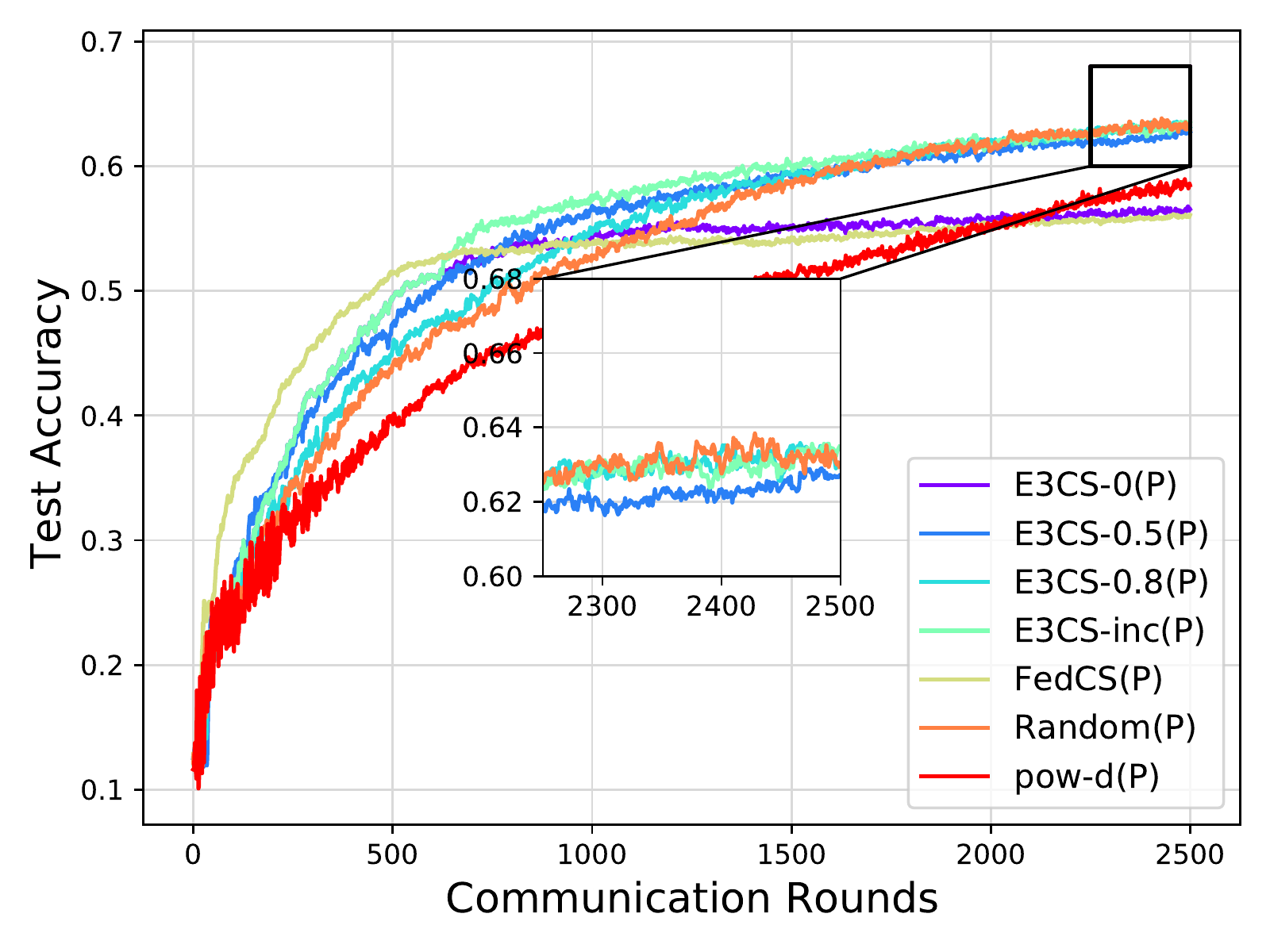}%
	}
	\caption{Test accuracy vs. communication rounds for CIFAR-10.}
	\label{acc CIFAR}
\end{figure*}
\begin{table*}[!htbp]
	\caption{Performance evaluation for CIFAR-10. Please note: 1)Data under Accuracy@\textbf{number} represents the first round to reach a certain test accuracy. 2) NaN means the accuracy never reaches the corresponding setting over the whole training session.}
	\label{ result CIFAR-10}
	\centering
	{\color{black}
	\begin{tabular}{ccc|cc|cc|cc}
	\midrule \multirow{2}{*}{Methods}       & \multicolumn{2}{c}{Accuracy@\textbf{45}}  & \multicolumn{2}{c}{Accuracy@\textbf{55}}  &  \multicolumn{2}{c}{Accuracy@\textbf{65}} &  \multicolumn{2}{c}{Final Accuracy (\%)}
		\\    \cmidrule(lr){2-3} \cmidrule(lr){4-5}  \cmidrule(lr){6-7} \cmidrule(lr){8-9}    & iid    &   non-iid      & iid    &  non-iid          & iid    & non-iid        & iid    &   non-iid  \\ 
		\midrule 
		E3CS-0(A)& 89& 307& 164& 636& NaN& NaN& 63.8& 61.0\\
		E3CS-0.5(A)& 106& 362& 198& 767& 468& 2252& 69.48& \textbf{65.6}\\
		E3CS-0.8(A)& 123& 429& 228& 849& \textbf{454}& 2279& 70.94& 65.42\\
		E3CS-inc(A)& 89& 307& 170& \textbf{631}& 650& \textbf{2127}& 71.12& 65.5\\
		FedCS(A)& \textbf{59}& \textbf{246}& \textbf{119}& 817& NaN& NaN& 60.6& 57.78\\
		Random(A)& 121& 472& 251& 930& 515& 2156& 71.46& 65.18\\
		pow-d(A)& 126& 689& 328& 1517& 742& NaN& \textbf{72.52}& 60.54\\
		\midrule
		E3CS-0(P)& 130& 374& 252& 1130& NaN& NaN& 64.18& 56.56\\
		E3CS-0.5(P)& 160& 406& 292& 863& 749& NaN& 68.98& 62.7\\
		E3CS-0.8(P)& 183& 499& 356& 995& 720& NaN& 69.62& 63.04\\
		E3CS-inc(P)& 130& 374& 252& \textbf{710}& \textbf{698}& NaN& 69.68& \textbf{63.48}\\
		FedCS(P)& \textbf{93}& \textbf{285}& \textbf{189}& 1822& NaN& NaN& 60.06& 56.08\\
		Random(P)& 201& 533& 392& 1155& 806& NaN& \textbf{70.54}& 62.94\\ 
		pow-d(P)& 208& 736& 483& 1928& 1083& NaN& 70.38& 58.48\\
		\midrule
	\end{tabular}
}
\end{table*}
We respectively ran 400 and 2500 communication rounds for EMNIST-L and CIFAR-10 datasets over different selection schemes in order to evaluate the convergence speed as well as the final model performance in the real training. Respectively for the two datasets,  we depict in Fig. \ref{acc EMNIST}, and Fig. \ref{acc CIFAR} the training performance of different schemes under both iid and non-iid scenario,  and we further present some statistical data in Table \ref{ result EMNIST-Letter}, and Table \ref{ result CIFAR-10}. Based on the experimental results, we derive the following observations:

 \textbf{Impact of CEP in the initial stage.} It is interesting to see that Cumulative Effective Participation (CEP)  has a conspicuous impact on convergence speed in the initial training stage. For the FedAvg-based solution,  we see that FedCS(A), which consistently chooses the clients with higher success rate, and is confirmed by Fig. \ref{selection record} to be the selection scheme with the highest CEP, obtains the fastest growth of accuracy during the first stage of training for both EMNIST-Letter and CIFAR-10 dataset. It is followed closely by E3CS-0(A) and E3CS-inc(A), which are also confirmed as having relatively high CEP in that stage. The gap of convergence speed over different schemes can also be observed in Table \ref{ result EMNIST-Letter} and \ref{ result CIFAR-10}, in which we found that E3CS-0(A), E3CS-inc(A), and FedCS(A) all has been accelerated to reach a certain fixed accuracy, compared with the vanilla selection scheme Random. {\color{black} By contrast, pow-d, which we confirm to have a relatively low CEP, does not promise us a commensurate convergence speed in our simulated volatile context. This may imply that when clients can drop out, the heuristic idea of always selecting the clients with higher loss might not necessarily accelerate convergence.}  Moreover, the impact on convergence speed seems to grow more significant in non-iid scenarios, as the gap between the "fastest" (i.e., FedCS(A)) and the "slowest" (i.e., pow-d(A)) has further expanded under this trend. Based on such an observation and since it is generally believed that non-iid data would further enhance the training difficulty, we conjecture that the difficulty of the task might have some sort of influence on CEP's impact on convergence speed, i.e., the more difficult the task is, the greater the influence of CEP will be. {\color{black}This conjecture is also aligned with another observation that the impact of CEP in accelerating the training is more significant for training on the CIFAR-10 dataset, a harder task compared with training on EMNIST-L. Similar observations can also be found for FedProx-based solutions, as depicted Fig. \ref{acc EMNIST} (c), (d) and Fig.\ref{acc CIFAR}(c),(d). }
 
 \textbf{Diminished impact of CEP.} However, the  effect of CEP diminishes in the middle/later stage of training. When the accuracy reaches a certain accuracy, aggregating more successful returns does not benefit much to the FL process, as we can observe that other fairer schemes, e.g., Random(A), though with a smaller CEP, gradually emulates FedCS(A) and eventually dominates it with the evolvement of the training process. A similar phenomenon is also observable for FedProx-based training. 
 
 \textbf{Impact of fairness.} The impact of the fairness factor is visually observable when training reaches its convergence. For all the experimental groups, we see that the final test accuracy of the most unfair scheme, FedCS, is the lowest among the evaluated methods. The method with the second-lowest accuracy is E3CS-0, which is also quite an "unfair" selection scheme, as it does not reserve probability for each client to ensure fairness.

{\color{black}\textbf{Motivation behind incremental fairness quota.}}
Based on the above observations, we find that i) CEP is critical for the initial stage of training in order to yield a faster convergence speed, but, ii) the effect diminishes with the training rounds goes, and that iii) the importance of fairness reinforces when the model approaches convergence. Our motivation to propose E3CS-inc is exactly based on the above observations. In E3CS-inc, we set $\sigma_t=0$ during the first $T/4$ rounds, so the algorithm will have no regard for fairness and put all the focus on increasing CEP. For the later $3T/4$ rounds, when the model has some sort of "over-fitting" on a portion of frequently selected data, we expand the selection fairness by making {\color{black} $\sigma_t=k/K$} (which makes it exactly an unbiased random selection), in order to make data on those seldom access clients being available for FL training. According to our result, E3CS-inc yields a performance as we have expected: it gets a very promising convergence speed in the first stage while its final test accuracy does not suffer an undesirable drop.

\textbf{Varying selection cardinality.}  We test different selection schemes based on varying selection cardinality $k=$10, 20, and 30. Due to the space limit, this part of content has been moved to Appendix \ref{varying selection}.
\par

\par

\section{Conclusion and Future Prospect}
In this paper, we have studied a joint optimization problem under the volatile training context. During our investigation,  we empirically discover a tradeoff between cumulative effective participation and fairness during the selection process, which in essence leads to another tradeoff between training convergence speed and final model accuracy. Aiming at optimizing the tradeoffs, we propose E3CS,  an efficient stochastic selection algorithm, for which we further design a practical setting of "fairness quota", such that the algorithm is enabled to tame the tradeoff between training convergence speed and final model accuracy. \par

{\color{black}In this paper, we propose to decompose the global problem P1 into two sub-problems based on the idea of alternating minimization. In our solution process, we focus on optimization of the client selection sub-problem (i.e., P1-SUB2) while fixing the solution of another sub-problem (i.e., P1-SUB1). However, as all the two essential components in FL, i.e., the local update operation and the client selection decision are mutually coupling, joint optimization of them needs to be further considered in future work.    }
%\section*{Acknowledgment}
%\small{
%This work is supported by National Natural Science Foundation of China (62072187, 61872084), Guangzhou Science and Technology Program key projects (202007040002, 201902010040, 201907010001), Guangdong Major Project of Basic and Applied Basic Research(2019B030302002), Key-Area Research and Development Program of Guangdong Province (2020B010164003), and the Fundamental Research Funds for the Central Universities, SCUT(Grant No. 2019ZD26).

\ifCLASSOPTIONcaptionsoff
  \newpage
\fi



%\bibliographystyle{IEEEtran}
\bibliographystyle{IEEEtran}
\bibliography{E3CS}{}
\newpage
\onecolumn
\appendices
{\color{black}
\section{Baseline Description}
\label{baseline description}
\begin{itemize}
	\item \textbf{FedCS.} FedCS is a client selection scheme originally proposed by Nishio \textit{et al.} in \cite{nishio2019client}, but the original proposal does not involve consideration of a volatile training context. As such, we make a minor revision to the algorithm in order to adapt it to our context.  The basic idea of FedCS is essentially to schedule clients that are faster in order to obtain more successful returns before the deadline. In our context, we assume that FedCS will consistently choose those clients with the highest success rate, and we allow FedCS to be prophetic, i.e., it has full information of the success rate of clients. 
	
		\item \textbf{Random.} Random is the vanilla \cite{mcmahan2016communication}, but also the most-accepted FL client selection scheme (applied both in FedAvg\cite{mcmahan2016communication} and FedProx\cite{li2018federated}). Random treats every client evenly that the server randomly selects a fraction of clients to participate in the training at each round. 
		\item \textbf{pow-d.} pow-d\footnote{\color{black}We add this method for comparison during our major revision in April, 2021.} is proposed in our concurrent work \cite{cho2020client}. In this work, selection fairness has also been highlighted, and its objective (barring the reservation of some degree of fairness) is to select the clients having larger local loss. Explicitly, pow-d randomly chooses a candidate set (with size $d$) and makes those in the set report their local loss on the current global model. Then it selects $k$ out of $d$ clients with the highest loss in the candidate set to formally start local training.    This work shares some common insights with ours (that bias selection may hurt performance), but the authors did not consider the potential dropout of clients. We like to compare this scheme with ours in the volatile training environment we set up. For a fair comparison, we assume that the clients always successfully report their local loss to the server, but are still vulnerable to dropout during training phase.  
	
\end{itemize}
}
\section{Proof of Regret }
\label{regret proof}
\subsection{Preliminary Facts }
We first prepare several facts that would be used in our proof. 
\begin{fact}[weight update, see Eq. (\ref{weight update final})]
 \label{fact1}
\begin{equation}
w_{i,t+1}= \begin{cases} w_{i,t} \exp\left(\frac{(k-K\sigma_{t})\eta \hat{x}_{i,t}  }{K} \right )  &  i \notin S_t \\ w_{i,t} &  i \in S_t   \end{cases}
 \end{equation}
\end{fact}
\begin{fact}[see Taylor series of exponential]
\label{fact2}
\begin{equation}\exp (x) \leq 1+x+x^{2} \quad  {  \text{\rm for all} } \quad  x \leq 1  \end{equation}
\end{fact}
\begin{fact}[see Taylor series of exponential]
\label{fact3}
\begin{equation} 1+x \leq \exp (x)  \quad  {  \text{\rm for all} } \quad x \in \mathbb{R}\end{equation}
\end{fact}
\begin{fact}[see probability allocation in Eq. (\ref{temp inter group selection probability})]
\label{fact4}
\begin{equation}  p_{i,t} - \sigma_t =  (k- K\sigma_t)\frac{ w_{i,t}}{\sum_{j \in \mathcal{K}} w_{j,t} }  \end{equation}
\end{fact}
\begin{fact}[see unbiased estimator in Eq. (\ref{unbised estimation final})]
\label{fact5}
{\color{black}
\begin{equation}
\hat{x}_{i,t}=\frac{\mathbb{I}_{ \{ i \in A_t\} }}{p_{i,t}}x_{i,t}
\end{equation}
}
\end{fact}
\begin{fact}[weighted AM-GM Inequality] \label{fact6} If $a_1, a_2, \dotsc, a_n$ are nonnegative real numbers, and $\lambda_1, \lambda_2, \dotsc, \lambda_n$ are nonnegative real numbers (the "weights") which sum to 1, then
\begin{equation}\sum_{i=1}^{n} \lambda_{i} a_{i} \geq \prod_{i=1}^{n} a_{i}^{\lambda_{i}}\end{equation}
\end{fact}

\begin{fact}[see $q_{i,t}^*$ in Definition \ref{optimal solution}]
 \label{fact7}
\begin{equation}
\sum_{i \in \mathcal{K}}q_{i,t}^*=1
 \end{equation}
\end{fact}

\begin{fact}\label{fact8} Since $k< K$,  $\hat{x}_{i,t}<1$ and $\eta<1$, we have:
\begin{equation} \frac{(k-K\sigma_{t})\eta \hat{x}_{i,t}  }{K} \leq 1 \end{equation}
\end{fact}

\begin{fact}\label{fact9} Combining Definition \ref{optimal solution} and $p_{i,t} \leq 1$  we have:
\begin{equation}  q_{i,t}^* \left (k-K\sigma_t \right) \leq 1-  \sigma_t  \end{equation}
\end{fact}

\subsection{Upper Bound of $  \mathbb{E}\left [\ln\frac{W_{t+1}}{W_t} \right ]$}
Now we formally start our proof. Let $W_t=\sum_{j=1}^K w_{j,t}$ and we first need to upper bound $\frac{W_{t+1}}{W_t}$. 
 Intuitively, we can derive the following inequality:
\begin{align*}
\frac{W_{t+1}}{W_t}=& \sum_{i \in \mathcal{K}-S_t} \frac{w_{i,t+1}}{W_{t}}+ \sum_{i \in S_t} \frac{w_{i,t+1}}{W_{t}}\\
 =&\sum_{i \in \mathcal{K}-S_t}  \frac{w_{i,t}\exp\left(\frac{(k-K\sigma_t)\eta \hat{x}_{i,t}  }{K}  \right) }{W_t}+ \sum_{i \in S_t} \frac{w_{i,t+1}}{W_{t}} \qquad \qquad \tag{\text{see   Fact  \ref{fact1}} } \\
 \leq&\sum_{i \in \mathcal{K}-S_t} \frac{w_{i,t} (1+\frac{(k-K\sigma_t)\eta \hat{x}_{i,t}  }{K} +\frac{(k-K\sigma_t)^2 \eta^2 \hat{x}_{i,t}^2  }{K^2})  }{W_t}+\sum_{i \in S_t} \frac{w_{i,t+1}}{W_{t}}  \qquad \qquad \tag{\text{combining   Fact \ref{fact2} and Fact   \ref{fact8}}} \\
=&1+\sum_{i \in \mathcal{K}-S_t}  \frac{w_{i,t} \left (\frac{(k-K\sigma_t)}{K} \eta \hat{x}_{i,t}+\frac{(k-K\sigma_t)^2}{K^2} \eta^2 \hat{x}^2_{i,t} \right  )}{W_t}  \qquad \qquad \tag{\text{since  $\sum_{i \in K} \frac{w_{i,t+1}}{W_{t}}=1$ }}  \\
 \leq&\exp \left (\sum_{i \in \mathcal{K}-S_t}  \frac{w_{i,t} \left(\frac{(k-K\sigma_t)}{K} \eta \hat{x}_{i,t}+\frac{(k-K\sigma_t)^2}{K^2} \eta^2 \hat{x}^2_{i,t} \right  )}{W_t} \right)  \tag{\text{see Fact \ref{fact3}}}\\
 =&\exp \left (\frac{1}{K} \sum_{i \in \mathcal{K}-S_t}  \left [\left (\frac{w_{i,t}(k-K\sigma_t)}{{W_t}}\right) \left( \eta \hat{x}_{i,t}+\frac{(k-K\sigma_t)}{K} \eta^2 \hat{x}^2_{i,t} \right ) \right] \right)\\
= & \exp \left (\frac{1}{K} \sum_{i \in \mathcal{K}-S_t}  \left [ (p_{i,t} - \sigma_t) \left( \eta \hat{x}_{i,t}+\frac{(k-K\sigma_t)}{K} \eta^2 \hat{x}^2_{i,t} \right ) \right] \right)  \qquad \qquad \tag{\text{see Fact \ref{fact4} }}      \\ 
\end{align*}
Taking the logarithm of both sides and summing using telescope, it yields:
\begin{align*}
\ln \frac{W_{T+1}}{W_1}&=  \sum_{t=1}^T \ln \frac{W_{t+1}}{W_t}    \\
&\leq \frac{1}{K}\sum_{t=1}^T  \sum_{i \in \mathcal{K}-S_t} \left( (p_{i,t}- \sigma_t) \eta \hat{x}_{i,t}+\frac{(k-K\sigma_t) ( p_{i,t} - \sigma_t)}{K} \eta^2 \hat{x}^2_{i,t} \right) \\
& \leq \frac{\eta}{K}\sum_{t=1}^T \sum_{i \in \mathcal{K}-S_t}  (p_{i,t}- \sigma_t)  \hat{x}_{i,t}+ \frac{ \eta^2}{K^2} \sum_{t=1}^T  \sum_{i \in \mathcal{K}} (k-K\sigma_t)   p_{i,t} \hat{x}^2_{i,t}   \\ 
\end{align*}
Taking expectation of the both sides, it yields:
\begin{equation}
\begin{split}
\mathbb{E} \left [\ln \frac{W_{T+1}}{W_1} \right ]
& \leq  \frac{\eta}{K}\sum_{t=1}^T \sum_{i \in \mathcal{K}-S_t} (p_{i,t}- \sigma_t)  \hat{x}_{i,t} +  \frac{\eta^2}{K^2}  \sum_{t=1}^T  (k-K\sigma_t)   \mathbb{E}\left [   \sum_{i \in \mathcal{K}} p_{i,t} \hat{x}^2_{i,t}  \right]
\end{split}
\end{equation}
Furthermore, we know that:
{\color{black}
\begin{align*}
\mathbb{E}\left [ \sum_{i \in \mathcal{K}}p_{i,t}   \hat{x}^2_{i,t} \right ]&\leq  \mathbb{E}\left [ \sum_{i \in \mathcal{K}} p_{i,t}\left  (   \frac{\mathbb{I}_{ \{ i \in A_t\} }}{p_{i,t}}x_{i,t} \right)^2  \right] \qquad \qquad \tag{\text{ see Fact \ref{fact5} }} \\
& \leq \mathbb{E} \left [ \sum_{i \in \mathcal{K}}  \frac{\mathbb{I}_{ \{ i \in A_t\} }}{p_{i,t}}x_{i,t}^2  \right]\\
& \leq \sum_{i \in \mathcal{K}} x_{i,t}^2   \\
& \leq K \tag{\text{ since $x_{i,t}\leq 1$ }}
\end{align*}}
It immediately follows that:
\begin{equation}
\label{right W}
\mathbb{E} \left [\ln \frac{W_{T+1}}{W_1} \right ] \leq  \frac{\eta}{K}\sum_{t=1}^T \sum_{i \in \mathcal{K}-S_t} (p_{i,t}- \sigma_t)  \hat{x}_{i,t} + \frac{\eta^2}{K} \sum_{t=1}^T  (k-K\sigma_t) 
\end{equation}
\subsection{Lower Bound of  $  \mathbb{E}\left [\ln\frac{W_{t+1}}{W_t} \right ]$}

On the other hand, we know that:
\begin{align*}
\ln \frac{W_{T+1}}{W_1}&=\ln \frac{  \sum_{i \in \mathcal{K}} w_{i,T+1} }{K}\\
&\geq \ln \frac{ \sum_{i \in \mathcal{K}}  q_{i,t}^* w_{i,T+1} }{K}  \tag{\text{ see Fact \ref{fact7} }}    \\
& \geq  \sum_{i \in \mathcal{K}}q_{i,t}^* \ln w_{i,T+1}  - \ln{{K}}   \tag{\text{combining Fact \ref{fact6} and  Fact \ref{fact7} }} \\
& \geq   \sum_{i \in \mathcal{K}-S_t}q_{i,t}^* \ln w_{i,T+1}  - \ln{{K}}   \\
&  \geq    \frac{\eta}{K}\sum_{t=1}^T \sum_{i \in \mathcal{K}-S_t}q_{i,t}^*(k-K\sigma_t)  \hat{x}_{i,t} - \ln{{K}}\tag{\text{see Fact \ref{fact1} }}
\end{align*}

Now taking expectation of both sides, we see that:
\begin{equation}
\begin{split}
\label{left W}
\mathbb{E}\left[\ln \frac{W_{T+1}}{W_1} \right] \geq   \frac{\eta}{K}\sum_{t=1}^T \sum_{i \in \mathcal{K}-S_t}q_{i,t}^*(k-K\sigma_t)  x_{i,t} - \ln{{K}}
\end{split}
\end{equation}
\subsection{Combining the bounds}

Combining Eqs. (\ref{right W}) and (\ref{left W}), finally we derive that:
\begin{equation}
\label{second to last inequality}
\sum_{t=1}^T \sum_{i \in \mathcal{K}-S_t}q_{i,t}^*(k-K\sigma_t)  x_{i,t} -  \frac{K}{\eta} \ln{{K}}  \leq   \sum_{t=1}^T \sum_{i \in \mathcal{K}-S_t} (p_{i,t}- \sigma_t)  \hat{x}_{i,t} +  \eta \sum_{t=1}^T     (k-K\sigma_t) 
\end{equation}
Moreover,  combining Fact \ref{fact9} and the fact that $1-\sigma_t =p_{i,t}-\sigma_t$ holds for $i \in S_t$ (as per the calculation of $p_{i,t}$, $p_{i,t}=1$ for all $i \in S_t$), trivially we can ensure: $  \sum_{ i  \in \mathcal{K} \cap S_t} q_{i,t}^* (k-K\sigma_t) x_{i,t} \leq  \sum_{i \in \mathcal{K} \cap S_t}  (p_{i,t}-\sigma_t)  x_{i,t} $. Plugging this fact into (\ref{second to last inequality}) , it yields:
\begin{equation}
\label{first plug}
 \sum_{t=1}^T \sum_{i \in  \mathcal{K}}  q_{i,t}^* (k-K\sigma_t)  x_{i,t}- \frac{K}{\eta} \ln{K}  \leq  \sum_{t=1}^T \sum_{i \in \mathcal{K}} (p_{i,t}- \sigma_t)  x_{i,t}+ \eta \sum_{t=1}^T    (k-K\sigma_t) 
\end{equation}
Reorganising the inequality, it yields:
\begin{equation}
 \sum_{t=1}^T \sum_{i \in  \mathcal{K}} \left( q_{i,t}^* \left (k-K\sigma_t \right)+ \sigma_t  \right)  x_{i,t}-  \sum_{t=1}^T \sum_{i \in \mathcal{K}} p_{i,t}  x_{i,t}  \leq \eta \sum_{t=1}^T    (k-K\sigma_t)  + \frac{K}{\eta} \ln{K}
\end{equation}
Plugging the definition of $ \mathbb{E} [\text{CEP}^*_T]$ and $\mathbb{E} [ \text{\rm CEP}^{\rm E3CS}_T ]$  (see  Eq. (\ref{Ecep*}) and Eq. (\ref{regret of E3CS}) ), we reach the following result:
\begin{equation}
R_T=\mathbb{E} [\text{CEP}^*_T]  -  \mathbb{E} [ \text{\rm CEP}^{\rm E3CS}_T ]    \leq  \eta \sum_{t=1}^T    (k-K\sigma_t)  + \frac{K}{\eta} \ln{K}
\end{equation}
Now we let $\eta= \sqrt{\frac{K \ln K}{ \sum_{t=1}^T     (k-K\sigma_t)}}$ and plug it into the above inequality, it yields:
\begin{equation}
R_T=\mathbb{E} [\text{CEP}^*_T]  -  \mathbb{E} [ \text{\rm CEP}^{\rm E3CS}_T ]  \leq  2\sqrt{ \sum_{t=1}^T K (k-K\sigma_t)      \ln K    } 
\end{equation}
This completes the proof.

\color{black}
\section{Additional Experiments under Varying Selection Cardinality }
\label{varying selection}
\begin{figure*}[!hbtp]
	\centering
	\subfloat[\color{black} iid, EMNIST-Letter ]{\includegraphics[width=1.8in]{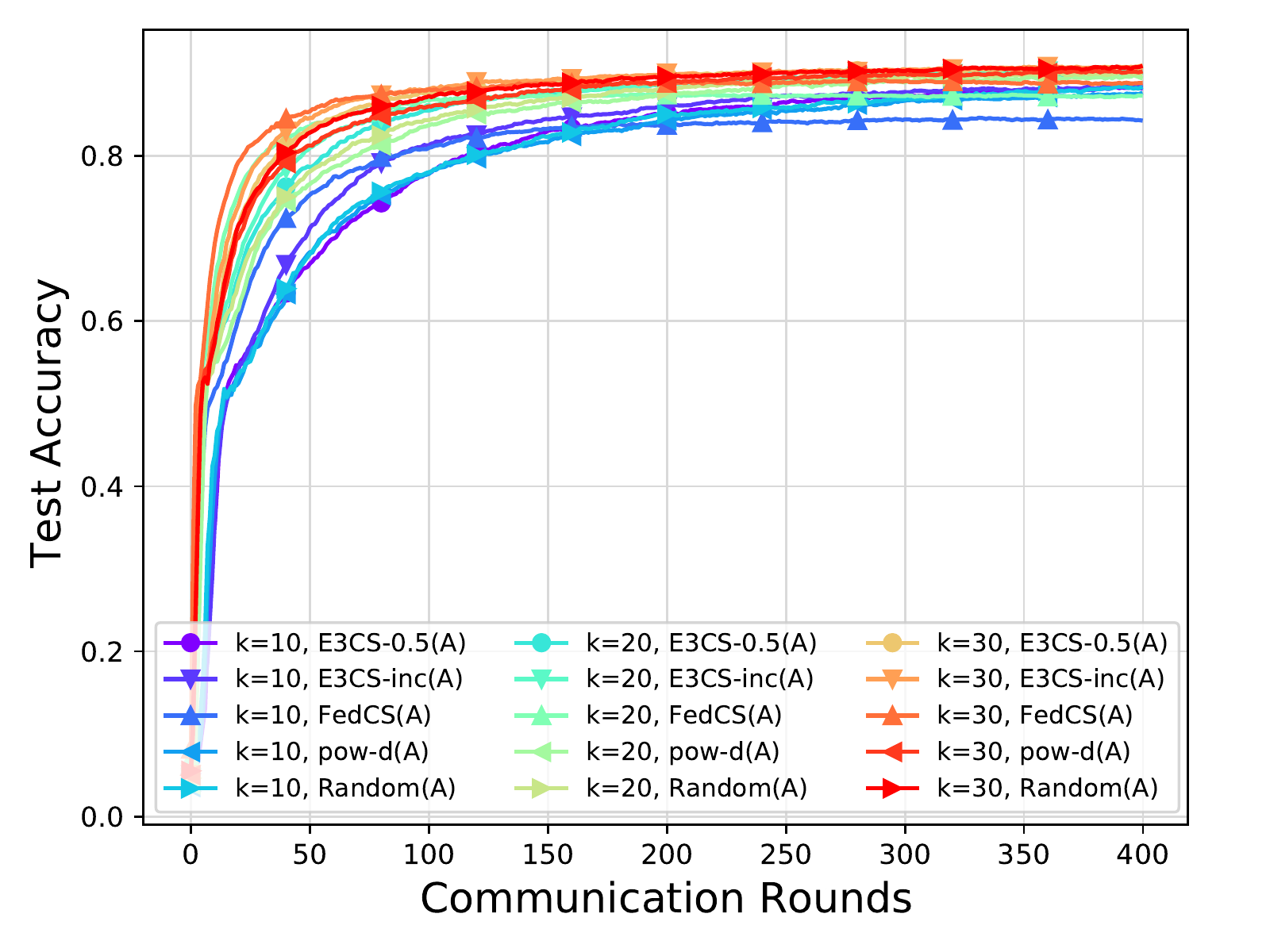}%
	}
	\subfloat[\color{black} iid, EMNIST-Letter ]{\includegraphics[width=1.8in]{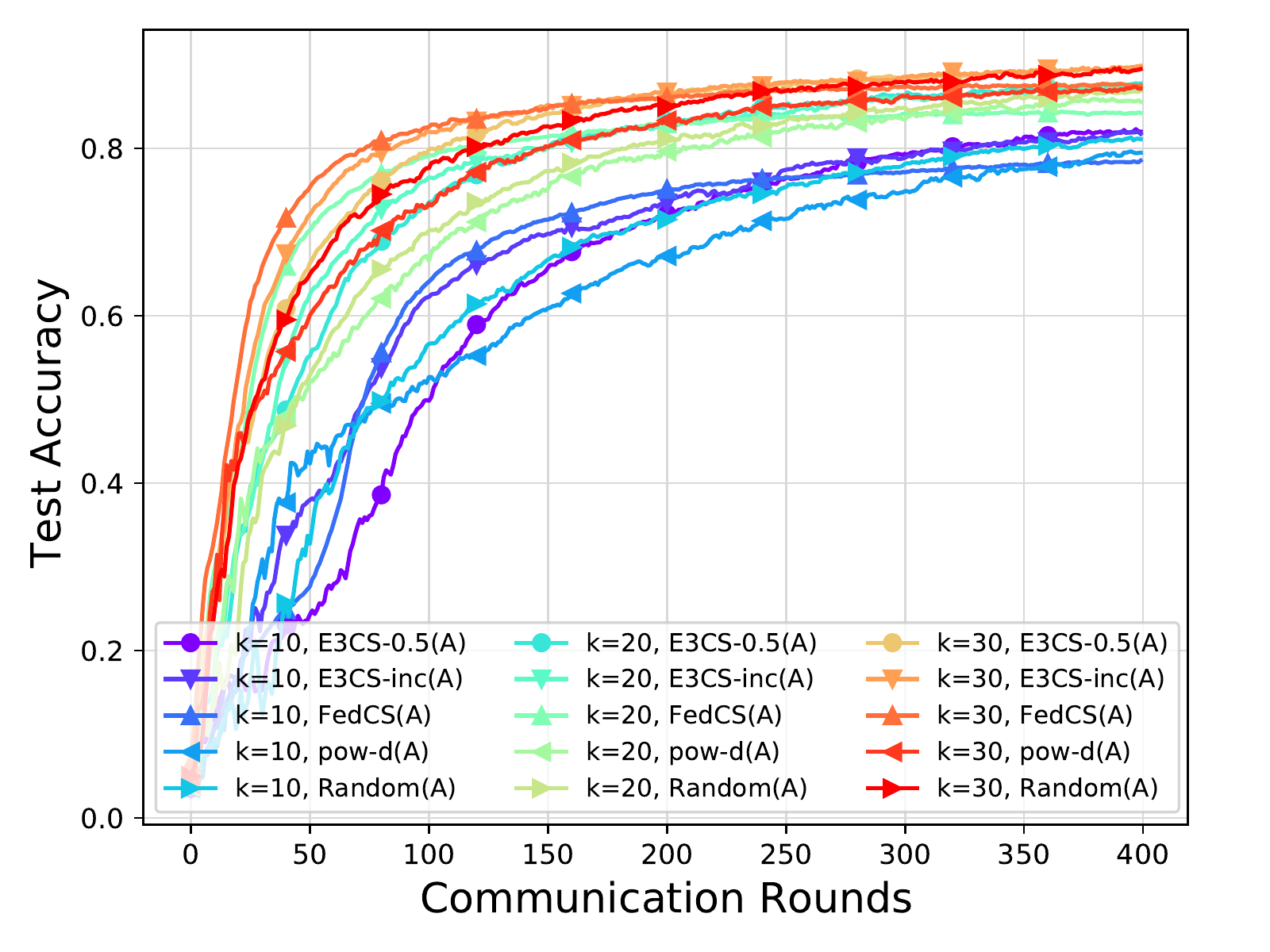}%
	}	
	\subfloat[\color{black} iid, CIFAR-10 ]{\includegraphics[width=1.8in]{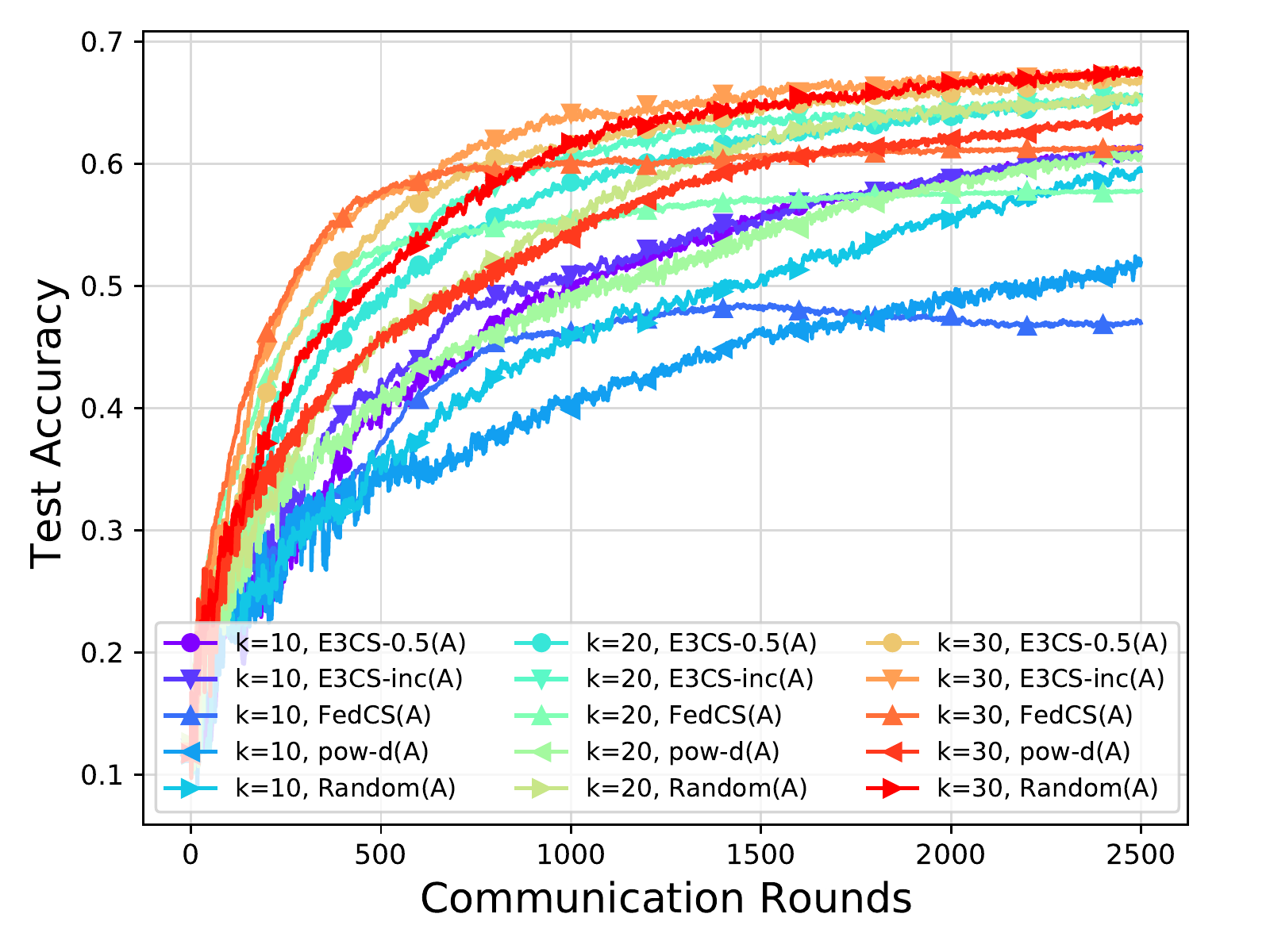}%
	}
	\subfloat[\color{black} non-iid, CIFAR-10 ]{\includegraphics[width=1.8in]{fig/fractionAcc_CIFAR-10_08_A.pdf}%	\label{acc a}
	}
	\caption{\color{black}Test accuracy vs. communication rounds under different selection cardinality (i.e., varing $k$). }
\label{varying k}
\end{figure*}

To show how the selection cardinality affects the FL performance, we compare different FedAvg-based schemes in the setting of $k=$10, 20, and 30. As shown in Fig \ref{varying k}, we see that the number of selections each round might have a significant effect on FL's convergence speed as well as the final convergence accuracy. Our experimental results concur with the basic finding from \cite{mcmahan2016communication}, that a greater extent of parallelism could indeed boost the training efficiency. Moreover, our results also support an accelerated convergence speed of our proposed E3CS solutions, with no (or only minor) final accuracy loss. 

\end{document}